# A Comprehensive Survey on Multi-hop Machine Reading Comprehension Approaches

Azade, Mohammadi

PhD Candidate, University of Isfahan, Isfahan, Iran, azade.mohammadi@eng.ui.ac.ir

Reza, Ramezani*

Assistant Professor, University of Isfahan, Isfahan, Iran, r.ramezani@eng.ui.ac.ir

Ahmad, Baraani

Professor of Computer Engineering, University of Isfahan, Isfahan, Iran, ahmadb@eng.ui.ac.ir

**Abstract**: Machine reading comprehension (MRC) is a long-standing topic in natural language processing (NLP). The MRC task aims to answer a question based on the given context. Recently studies focus on multi-hop MRC which is a more challenging extension of MRC, which to answer a question some disjoint pieces of information across the context are required. Due to the complexity and importance of multi-hop MRC, a large number of studies have been focused on this topic in recent years, therefore, it is necessary and worth reviewing the related literature. This study aims to investigate recent advances in the multi-hop MRC approaches based on 31 studies from 2018 to 2022. In this regard, first, the multi-hop MRC problem definition will be introduced, then 31 models will be reviewed in detail with a strong focus on their multi-hop aspects. They also will be categorized based on their main techniques. Finally, a fine-grain comprehensive comparison of the models and techniques will be presented.

CCS CONCEPTS • Information Systems → Machine Reading comprehension • Computing Methodology → Machine Reading comprehension

**Additional Keywords:** Multi-hop Machine Reading Comprehension, Natural Language Processing

**Declarations:**

Funding: No

Conflicts of interest/Competing interests: No

Availability of data and material: Not applicable

Code availability: Not applicable

## 1 INTRODUCTION

Machine reading comprehension (MRC) is one of the most important and long-standing topics in the Natural Language Machine reading comprehension (MRC) is one of the most important and long-standing topics in Natural Language Processing (NLP). MRC provides a way to evaluate an NLP system's capability for natural language understanding. An MRC task, in brief, refers to the ability of a computer to read and understand natural language context and then find the answer to questions about that context. The emergence of large-scale single-document MRC datasets, such as SQuAD [1], CNN/Daily mail [2], has led to increased attention to this topic and different models have been proposed to address the MRC problem, such as (Chen, Bolton, and Manning, 2016) [4][5][6].

However, for many of these datasets, it has been found that models don't need to comprehend and reason to answer a question. For example, Khasabi et al. [7] proved that adversarial perturbation in candidate answers has a negative effect on the performance of the QA systems. Similarly, Jia and Liang [8] showed that adding an adversarial sentence to the SQuAD [1] context will drop the result of many existing models. Also Chen, Bolton, and Manning [3] pointed out that the required reasoning in the CNN/Daily Mail [2] dataset is so simple that even a relatively simple algorithm can perform well on this dataset. Min et al. [9] have shown that 90% of the questions in SQuAD [1], are answerable given only one sentence in a document then models usually can find the answers in a sentence that is matched with the question, which does not involve complex reasoning.

The above problems seem to be due to the fact that answering the questions of these datasets doesn't require a deep understanding and reasoning [10]. In other words, these models have focused only on answering questions based on a single or few nearby sentences of the context, mostly by matching information in the question and the context (known as single-hop MRC). However, in real-world cases, to answer a question it is required to read and comprehend multiple parts of disjoint evidence to find the valid information. Therefore, there are gaps between single-hop datasets and their models, and real-world applications. To eliminate these gaps, the single-hop MRC models are faced with three important challenges:

**Reasonability**: Since in real-world cases, to answer a question it is necessary to integrate and synthesize information from multiple pieces of information, more complex reasoning named multi-hop reasoning is required. Multi-hop reasoning means reasoning over information taken from more than one document [11].

**Interpretability**: Interpretability is a vital feature of any reliable system. Since in real-world cases, the evidence used to find the answer is not necessarily located closely and could be comprehended from disjointed pieces of information Then, finding all the supporting facts is more difficult [12].

**Scalability**: In real-world cases, for each question, there may be multiple supporting documents sets but only a small part of them contains valid information. Then scalability became a serious challenge which means that the models could be able to handle the increasing amount of input with a limited cost. Most single-hop models do not scale well because they have been built for one passage.

Multi-Hop Reading Comprehension (MHRC) is a more challenging extension of MRC in which the models need to properly integrate multiple pieces of evidence and reason over them to correctly answer a question. In contrast to the question in single-hop MRC that can be answered by matching information in some nearby sentences, multi-hop MRC requires answering more complex questions based on a deep understanding of the full information. The multi-hop reasoning (also known as multi-step reasoning) is considered the basic key in multi-hop MRC. Multi-hop reasoning is the ability of reach from some intermediate steps to a final step through a reasoning chain [13].

Figure 1 shows an example of Multi-hop MRC problem from HotpotQA [11]. In this case, the answer span *Greenwich Village, New York City* could be found in a shortcut as it is closely related to the span *New York City*. Also, finding evidence paragraph *Adriana Trigiani* requires the information from paragraph *Big Stone Gap (film)* [14].

Adriana Trigiani is an Italian American best-selling author of sixteen books, television writer, film director, and entrepreneur based in Greenwich Village, New York City.

(2) Trigiani has published a novel a year since 2000.

Paragraph 2, Big Stone Gap (film):

(1) Big Stone Gap is a 2014 American drama romantic comedy film written and directed by Adriana Trigiani and produced by Donna Gigliotti for Altar Identity Studios, a subsidiary of Media Society.

(2) Based on Trigiani's 2000 best-selling novel of the same name, the story is set in the actual Virginia town of Big Stone Gap circa 1970s.

(3) The film had its world premiere at the Virginia Film Festival on November 6, 2014.

**Paragraph 3,** …

**Question**: The director of the romantic comedy "Big Stone Gap" is based in what New York city?

**Answer**: Greenwich Village, New York City

**Evidence Sentences**: ["Big Stone Gap (film)", 1], ["Adriana Trigiani", 1]

Figure 1: An example of multi-hop MRC from HotPotQA

The first attempt to improve the simple single-hop MRC task happened with emerging of some datasets like TriviaQA [15] and NarrativeQA [16]. These datasets addressed more challenges by introducing multiple passages per each question and also presenting a more complex kind of questions that couldn't be answered with one single sentence. Although this kind of question was more complex than single-hop questions, they still could be answered by a few nearby sentences within one passage, which means they mostly do not need multi-hop reasoning. They are generally known as the multi-passage or multi-document dataset that is closer to open-domain Question Answering or retrieving-reading problems, which means models have to focus on retrieving the most related passage and then answer the question based on that passage instead of reasoning over information from multiple passages. HotpotQA [11] and WikiHop [17] can be mentioned as the first and most popular multi-hop datasets which in addition to providing multiple passages per each question, ensure that the question can only be answered by reasoning over disjoint pieces of information across different passages. It has been shown that the models with successful results in single-hop MRC datasets have limited success on these datasets [18].

Recently, a lot of studies have been done in the field of multi-hop MRC, they focus on different aspects of the task. One of the most basic aspects is to propose a model for solving the multi-hop MRC problem, which has received great attention in recent years. Due to the importance of this task, and also the high speed of presenting new models, it is necessary to present a comprehensive investigation of current models. It would clarify the advantages and disadvantages of existing solutions and help improve future models.

To have an accurate view of the growing trend of multi-hop MRC, Figure 2 has been prepared. In 2017, several datasets were introduced (like TriviaQA [15]) that are not considered multi-hop datasets, but they attracted attention to this task by proposing more complicated questions than single-hop questions. Although proposing multi-hop models has been beginning from 2018, there was a serious shortage of multi-hop datasets, so in 2018 some MRC datasets were been proposed with a main focus on the multi-hop challenges. These datasets made a proper situation to present the multi-hop MRC models, and as you can see a significant number of multi-hop models were been proposed in 2019. The trend of proposing new multi-hop datasets and models has been continued in 2020. It can be concluded from Figure 2 that after proposing each new dataset, some models will be proposed as well to address the new challenges of that dataset, as you can see the number of models in 2019 because of the large number of datasets in 2018. Besides, the small number of researches in 2022 is not due to the lack of attention to this field, but it is because the studies in 2022 have not yet been fully published.

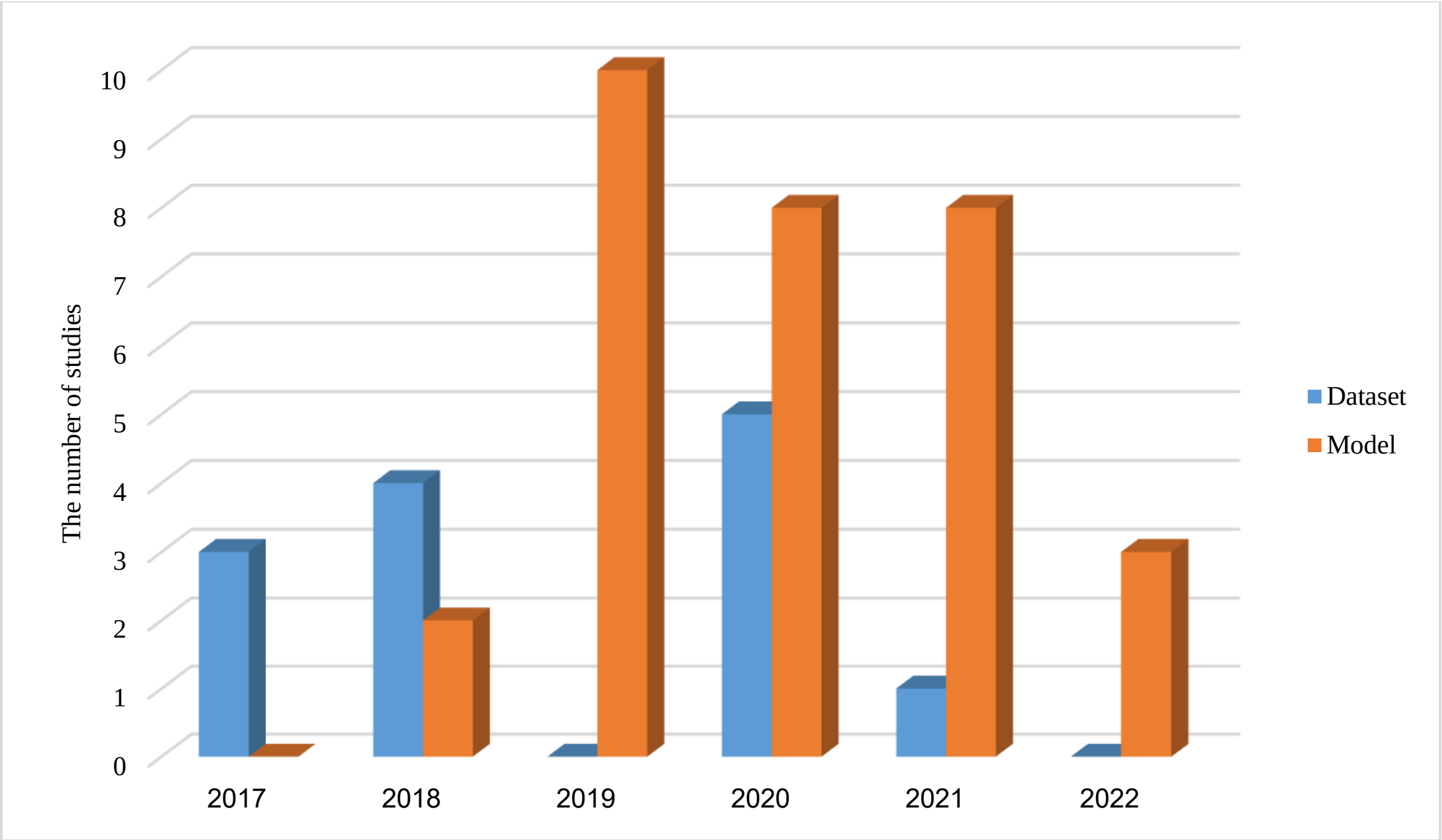

Figure 2: The frequency of the multi-hop models and datasets

There are some review papers on the MRC task that we will mention here to explain the difference and the innovation aspect of our paper. Liu et al. [19] reviewed 85 studies from 2015 to 2018 with a focus on neural network solutions for the MRC problem to investigate the neural network methods in the MRC task. Baradaran, Ghiasi, and Amirkhani [20] presented a survey on the MRC field based on 124 reviewed papers from 2016 to 2018 with a focus on presenting a comprehensive survey on different aspects of machine reading comprehension systems, including their approaches, structures, input/outputs, and research novelties. Thayaparan, Valentino, and Freitas [20] proposed a systematic review of explainable MRC, from 2014 to 2020 with a focus on the explainable feature of the recent MRC methods. Zhang, Zhao, and Wang [22] presented a survey on the role of contextualized language models (CLMs) on MRC from 2015 to 2019. Bai and Wang [23] presented a survey on textual question answering with a focus on datasets and metrics, they investigate 47 datasets and 8 metrics.

Although the above studies investigate different aspects of MRC/QA, none of them have focused on the multi-hop challenges. Due to the importance and increasing attention to multi-hop MRC it is necessary to investigate the multi-hop MRC studies separately. Our contribution in this paper is to focus on the multi-hop MRC models and techniques. In this regard, we first proposed the problem definition of the multi-hop MRC task, then categorize 31 models from 2018 to 2022 based on their main techniques and also investigate each model in detail. Also, a comprehensive comparison of the models and techniques will be presented. Finally, open issues in this field have been discussed.

It is important to note that since there is a close relationship between MRC and Question Answering, most of the existing machine reading comprehension tasks are in the form of textual question answering [24], also MRC is known as a basic task of textual question answering [19]. Thus, we consider cloze domain textual question answering as a typical MRC task in this paper. Figure 3 shows the relationship between QA and MRC and multi-hop MRC [23].

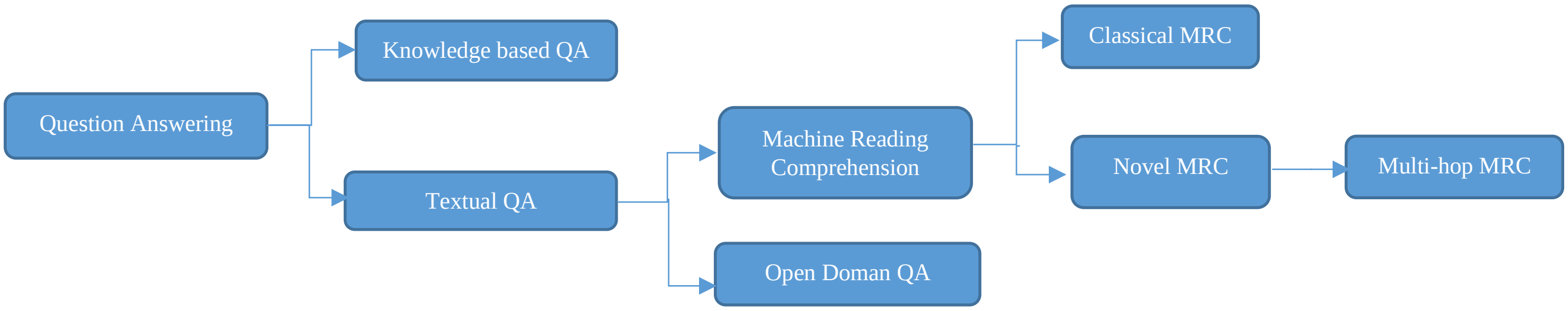

Figure 3: The relationship between QA and MRC and multi-hop MRC [23]

Our resources were Google, Google Scholar, IEEE Explore, and Elsevier. We searched for papers with these keywords: "multi-hop machine reading comprehension", "multi-hop question answering", "reading comprehension over multiple documents", "reading comprehension over multiple passages", "question answering over multiple documents", and "question answering over multiple passages".

The rest of this paper is organized as follows: In Section 2, the definition of the multi-hop MRC problem is explained. In section 3, besides categorizing 31 models based on their main techniques, each model will be reviewed in detail with a focus on its main idea and superiority. Next, a comprehensive comparison of the models and techniques will be done by presenting some figures and tables in section 4. Then, open issues are expressed in Section 5, and finally, Section 6 concludes the paper.

## 2 PROBLEM DEFINITION

In general, the multi-hop MRC problem can be defined as:

Given a collection of training examples $(C; Q; A)$, the goal is to find a function $F$ which takes a context $C$ and a corresponding question $Q$ as inputs, and gives answer $A$ as output.

$$F: (C, Q) \rightarrow A \qquad (1)$$

For the multi-hop MRC problem, $C = (P_1, P_2, \dots, P_{lp})$ can be a set of paragraphs (or documents) where $l_p$ denotes the number of paragraphs (or documents). Question $Q$ is such a way that needs the multiple disjoint pieces of information from $C$ to be answered. In other words, it needs multi-hop reasoning, each intermediate step of the reasoning chain can be considered a hop.

$$h = [s_1, s_2, \dots . s_n] \qquad n \geq 2 \qquad (2)$$

where $h$ is for hop and $s_i$ is the $i^{th}$ intermediate step. The number of hops have to be more or equal to 2.

Like general MRC task, Answer $A$ in multi-hop MRC can be in different forms, where have been divided into four categories[25]:

**Span-extraction**: The span extraction task needs to extract the subsequence $A$ from $P_i (P_i \in C)$ as the correct answer of question $Q$ by learning the function $F$, such that $A = F(C, Q)$.

**Multiple-choice**: Given a set of candidate answers $A = \{A_1, A_2, \dots, A_n\}$, the multiple-choice task needs to select the correct answer $A_i$ from $n$ possible answer by learning the function $F$, such that $A_i = F(C, Q)$.

**Free-form**: The correct answer is $A$ that $A \subseteq C \; or \; A \nsubseteq C$. In other words, the answer is not necessarily limited to be a part of the passage. The free-form task needs to predict the correct answer $A$ by learning the function $F$, such that $A = F(C, Q)$.

**Cloze-style**: The correct answer $A$ is part of the question $Q$ (usually a word or an entity) that is removed from question. The cloze style task needs to fill in the blank with the correct word or entity $A$ by learning the function $F$, such that $A = F(C, Q)$.

To show the frequency of each task among multi-hop studies Figure 4 has been prepared. This figure contains some points:

- Most studies have focused on the Span-extraction and Multiple-choice tasks. One of the main reasons is that most available MRC datasets are also in these forms and automatically this fact encourages research to focus on these two tasks.
- A few studies have focused on the free-form task; besides the lack of a proper datasets, it is also due to the complexity of this task.
- None of the studies have focused on the Cloze-style task. This task has potential to be used in multi-hop MRC as well as general MRC but there is no any Cloze-style multi-hop MRC dataset.

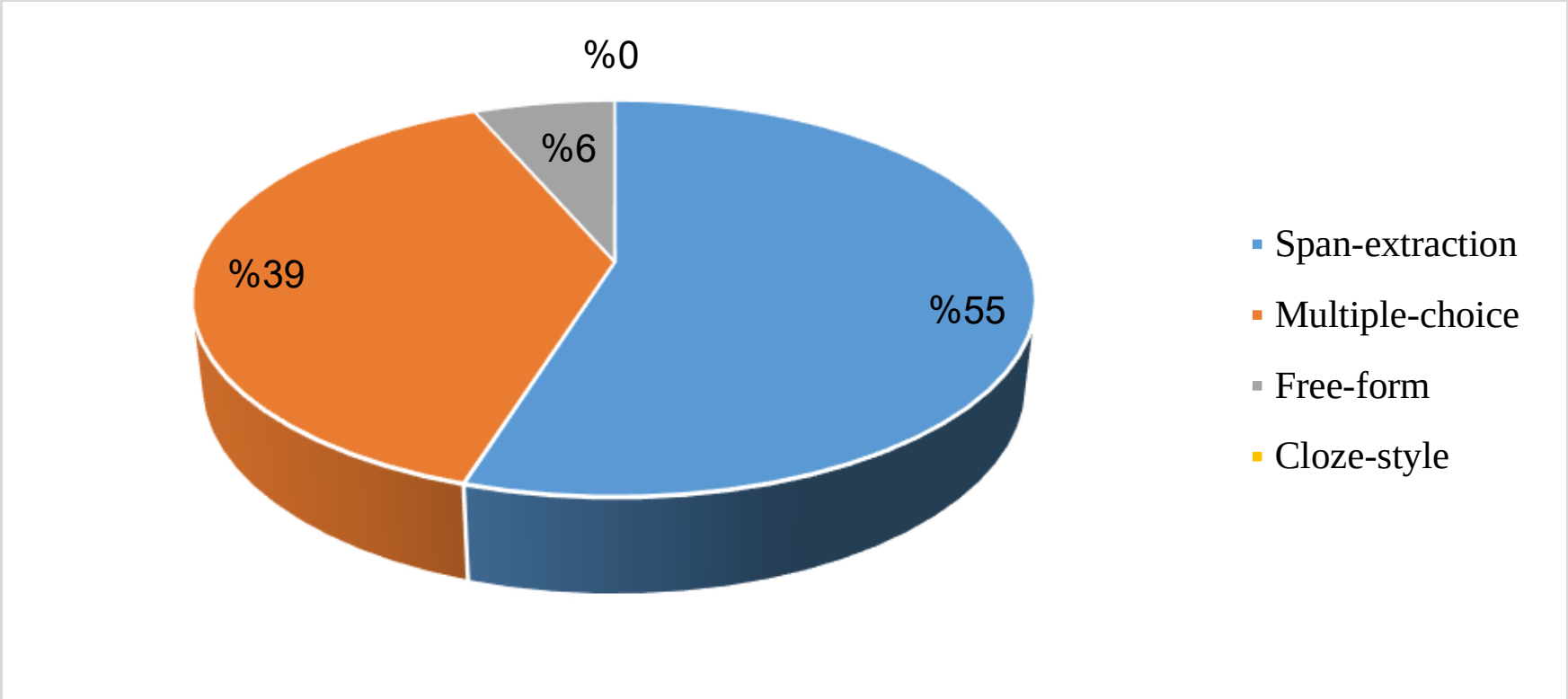

Figure 4: The frequency of Multi-hop MRC tasks in reviewed studies

## 3 MULTI-HOP MRC TECHNIQUES

In this paper, 31 studies have been investigated, which propose a model for multi-hop MRC based on the presented problem in section 2. It is important to note that since there is a close relationship between MRC and Question Answering, most of the existing machine reading comprehension tasks are in the form of textual question answering [24], and also MRC is known as a basic task of textual question answering [19]. Thus, we consider cloze domain textual question answering as a typical MRC task in this paper.

Existing studies for multi-hop MRC can be mainly divided into three categories: decomposition, recurrent reasoning based on memory retrieval, and multi-step reasoning based on graph neural networks. For each category, after explaining the main technique, the multi-hop MRC models will be reviewed in detail; beside reviewing the detail architectures of each model, we also focus on the superiority and the motivation of them. Also, the disadvantages of each technique will be discussed at the end. In the next section (4) a comprehensive comparison of the techniques and models will be presented.

The techniques do not have a specific order, because all three techniques have been used by models from 2018 to 2022 (Figure 40), but as much as possible, the studies within the techniques have been according to the order of published time.

### 3.1 Decomposition technique:

Complicated question is a basic challenge of multi-hop MRC, unlike the single-hop questions, they cannot be answered easily and require complicated reasoning. Since the human reasoning about complex questions is done by decomposition, answering sub-questions, summarizing, and comparing [26], then this technique has focused on simplifying the problem by decomposition of a complex question into multiple simple sub-questions. It means it reduces multi-hop MRC to multiple single-hop MRC. This technique mostly uses the single-hop MRC models to find the answers to sub-questions and then combine the answers to obtain the final answer. In the following, the models which use this technique for multi-hop MRC will be reviewed in detail.

**Self-assembling MNM**: Jiang and Bansal [18] focused on identifying the sub-questions in the correct reasoning order and presented an interpretable and controller-based self-assembling neural modular network for the multi-hop reasoning process which includes two main parts, Modular Network with a Controller (top) and the Dynamically-assembled Modular Network (bottom) that can be seen in Figure 10. The main idea of the model to handle multi-hop questions is done with Controller that computes an attention distribution over all question words at every reasoning step, which finds the sub-question that should be answered at the current step. In summary, Controller reads the question and predicts a series of modules that could be executed in order to answer the given question. Each module deals with a single-hop sub-question, then they will be chained together according to the predicated order by controller to get the final answer. The mentioned modules are described as follows: All modules take the question representation $u$, context representation $h$, and sub-question vector $ct$ as input.

- The *Find* module first builds a similarity matrix between the question and context, and then uses it to generate a question-aware context representation. Finally, this bi-attention result is pushed into a stack.
- The *Relocate* module pops an attention map from the stack and computes the bridge entity's representation, which is a weighted average over context representation. The representation is then used to compute a bridge-entity-aware representation of the context. Finally, it applies the Find module between the representations of the context and question.

- The *Compare* module pops two attention maps from the stack and computes two weighted averages over the context representation using the attention maps.
- The *NoOp* module can be seen as a skip command. It is used to reduce computations when the controller decides no action is required.

However, this system approaches question decomposition by having a decomposer model trained via human labels (Cao and Liu, 2021a).

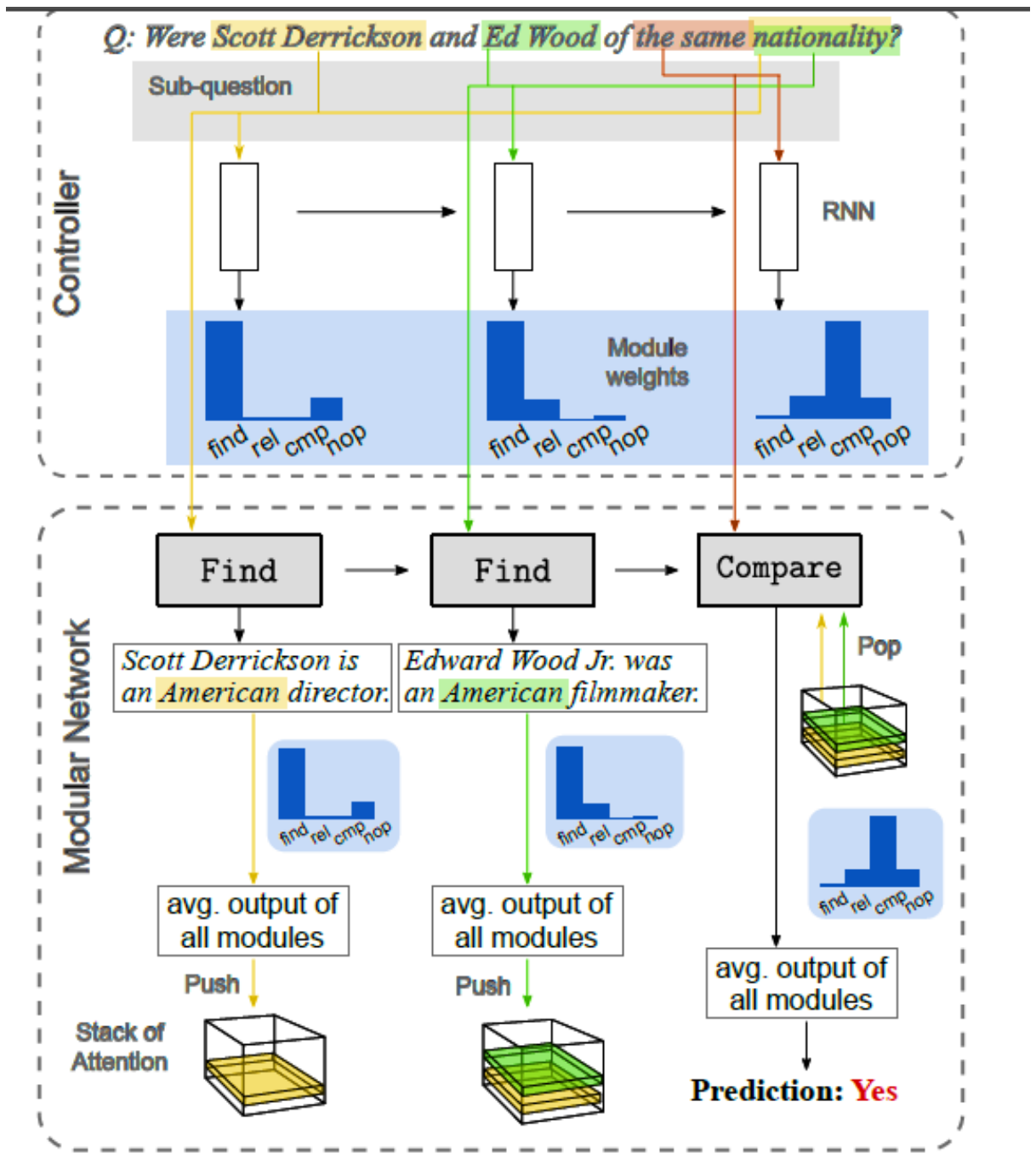

Figure 5: Architecture of Self-assembling MNM [18]

**ONUS**: Perez et al. [27] proposed an algorithm for One-to-N Unsupervised Sequence transduction (ONUS) that map a complicate multi-hop question to some simpler single-hop sub-questions. Unlike other decomposition studies use a combination of hand-crafted heuristics, rule-based algorithms, and learning from supervised decompositions to decompose multi-hop question which require significant human effort, this model automatically learns to decompose different kinds of questions. The main idea has been shown in Figure 6. To decompose multi-hop question $Q$ to simpler corpus $D$, First some candidate sub-question from a simple corpus $S$ will be created by mining 10M possible sub-questions from Common Crawl with a classifier. It then trains a decomposition model on the mined data using Q and D with unsupervised sequence-to-sequence learning to map multi-hop questions to sub-question. With this idea, the model is able to train a large transformer model to generate decompositions and avoid heuristic/extractive decompositions.

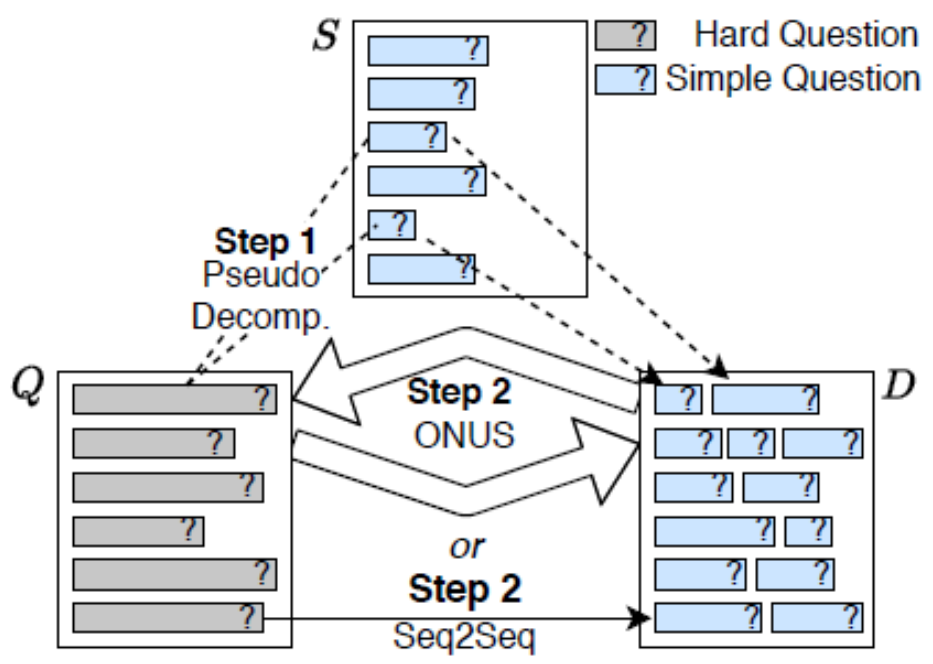

Figure 6: Overall process of ONUS [27]

**CGDe-FGIn**: Coa and Liu [12] proposed a Coarse-Grained Decomposition Fine-Grained Interaction (CGDe-FGIn) model to handle the complicated multi-hop questions. Existing studies use Bi-directional Attention Flow (Bi-DAF) to capture semantic feature interaction between documents and questions, but Bi-DAF generally captures the surface semantics of words, instead of the implied semantic feature of intermediate answers, so it cannot extract the most important parts of multiple documents in the multi-hop MRC task. CGDe-FGIn consists of:

- *Coarse-Grain complex question Decomposition (CGDe)* to decompose complex questions into simple sub-questions without any additional human annotations (Figure 7). A semantic similarity softmax matrix is calculated for the question and multiple documents according to the query direction.
- *Fine-Grained Interaction (FGIn)* to better represent each word in the document and extract more comprehensive and accurate sentences related to the inference path instead of using Bi-DAF (Figure 8). This layer generates the Context2Query attention to show which question words are most relevant to each context word.

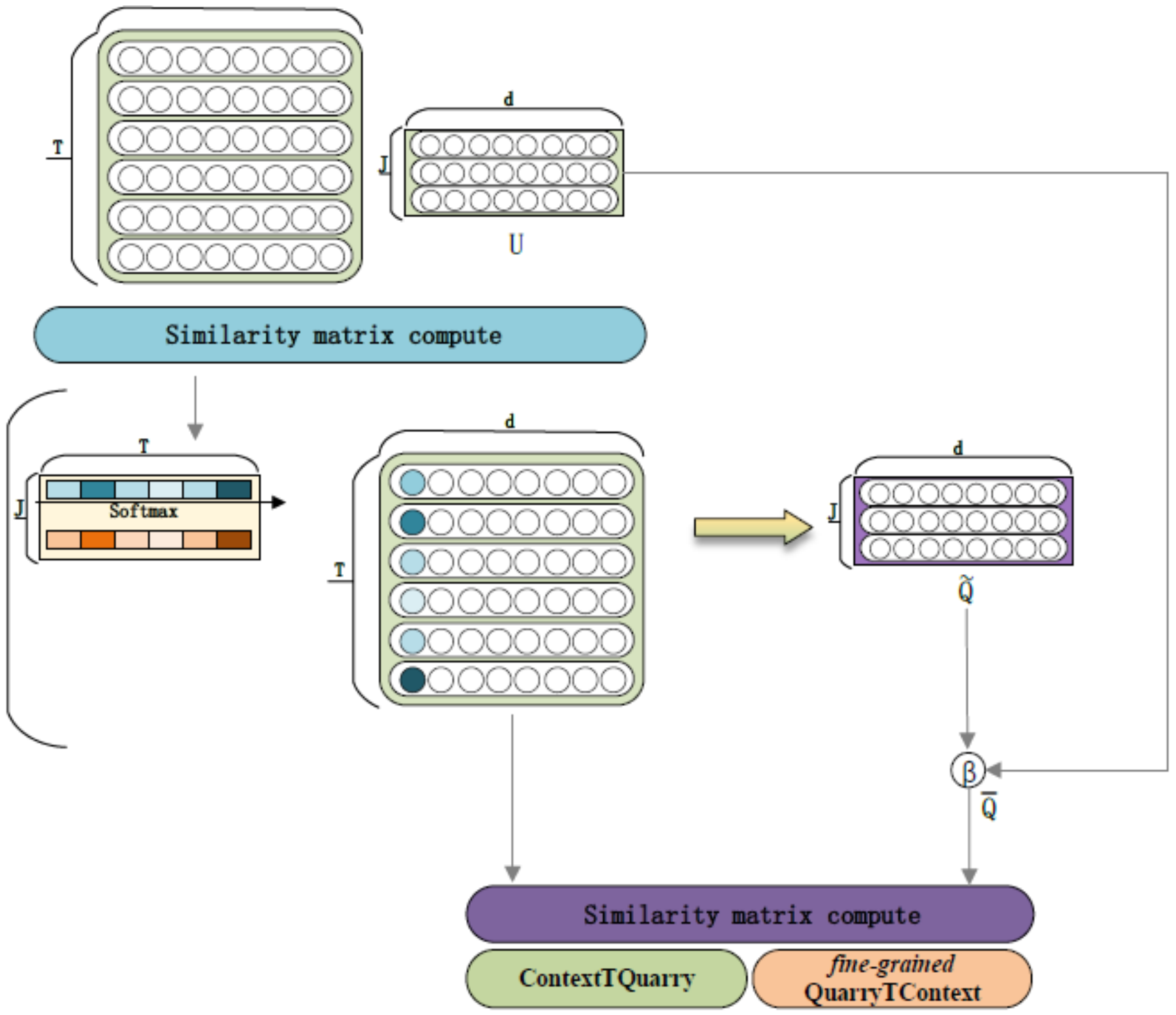

Figure 7: Coarse-grained decomposition [12]

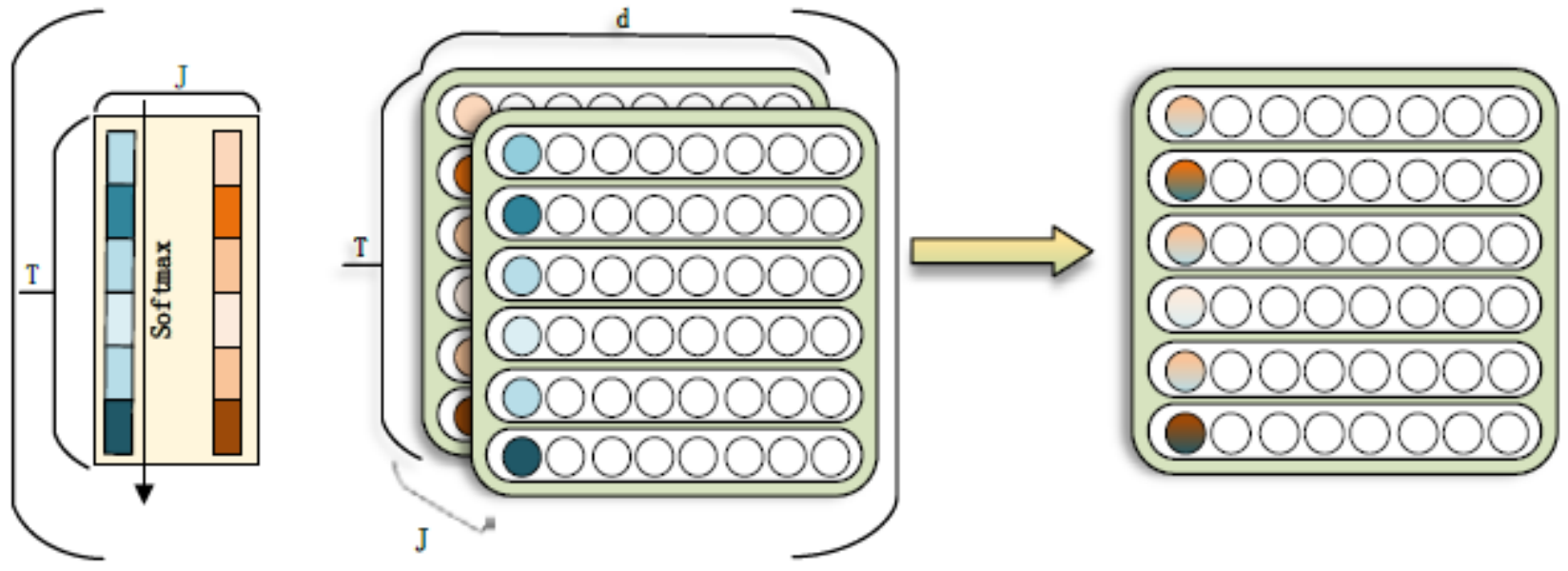

Figure 8: Fine-grained interaction [12]

**RERC**: Fu et al. [26] also handled multi-hop MRC with a focus on decomposing complicated questions into simpler ones. They proposed a three-stage Relation Extractor-Reader and Comparator (RERC) model. RERC is consist of 1) *Relation Extractor* to decompose the complex questions into simple sub-questions by automatically extracting the subject and key relations of the complex question, 2*) Reader* to find the answers to the sub-questions in turn by an advanced ALBERT model, and finally, 3) *Comparator* to perform the numerical comparison and summarizes all the answers to get the final answer (Figure 9)

The main part of this model is Relation Extractor with two different structures: 1) classification-type (CRERC), where the evidence relation information in the dataset is used as prior knowledge, and the question text is mapped to question relations through the classifier; 2) span-type (SRERC), where the type of question relations is unrestricted, and the Relation Extractor can automatically extract multiple corresponding spans from the question text as question relations.

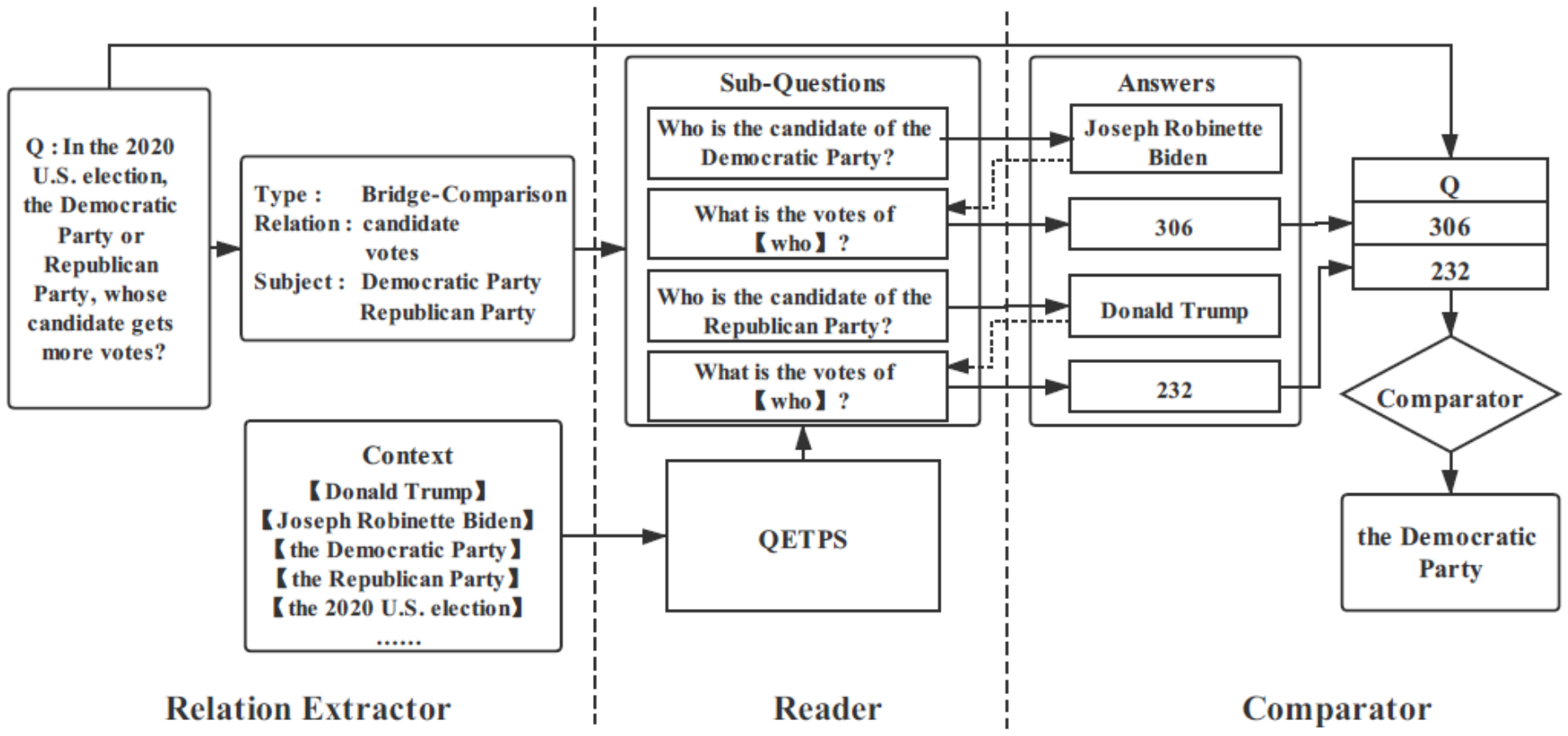

Figure 9: Architecture of the RERC models [26]

The *Decomposition* technique was one of the first ideas for multi-hop MRC and as you can see in recent years (2021) still has attracted attention. The main disadvantage of this technique is that, instead of focusing on multi-hop reasoning as an important key of multi-hop MRC, it focuses on reducing the problem to a single-hop MRC. Thus, they actually do not go far beyond single-hop models.

### 3.2 Recurrent reasoning-based technique

The sequence models have been first used for single-hop MRC tasks, and most of them are based on Recurrent Neural Networks (RNNs), some studies focus on using them in multi-hop MRC. It can be called state-based reasoning models and they are closer to a standard attention-based RC model with an additional "state" representation that is iteratively updated. The changing state representation results in the model focusing on different parts of the passage during each iteration, allowing it to combine information from different parts of the passage [28]. Most models, presented in this section, have used advanced neural network concepts, such as attention mechanism and network memory for multi-hop reasoning. In the following the models which use this technique will be reviewed in detail including the architecture alongside the superiority and the motivation of them.

**MHPGM**: Bauer et al. [29] found that 11% of the question in the Wikihop dataset (Song et al., 2020) and 42% of the question in the NarrativeQA dataset [16] require commonsense knowledge, then they proposed a two-stages model as has been shown in Figure 15: *Multi-Hop Pointer-Generator* Baseline that uses multiple hops of bidirectional attention, self-attention, and a pointer-generator decoder for multi-hop reasoning over the embedded context using k-reasoning cells, then a *Commonsense Algorithm* to select and insert grounded multi-hop relational knowledge paths from ConceptNet between the hops of document-context reasoning, via the Necessary and Optional Information Cell (NOIC). In other words, besides focusing on multi-hop reasoning using state-based reasoning, they also insert the grounded multi-hop relational knowledge from ConceptNet between the hops of document-context reasoning.

*Commonsense Algorithm* consists of *Commonsense Selection Representation* to select useful relational knowledge paths and *Commonsense Model Incorporation* to fill the gaps of reasoning between hops of inference using *Necessary and Optional Information Cell (NOIC)*.

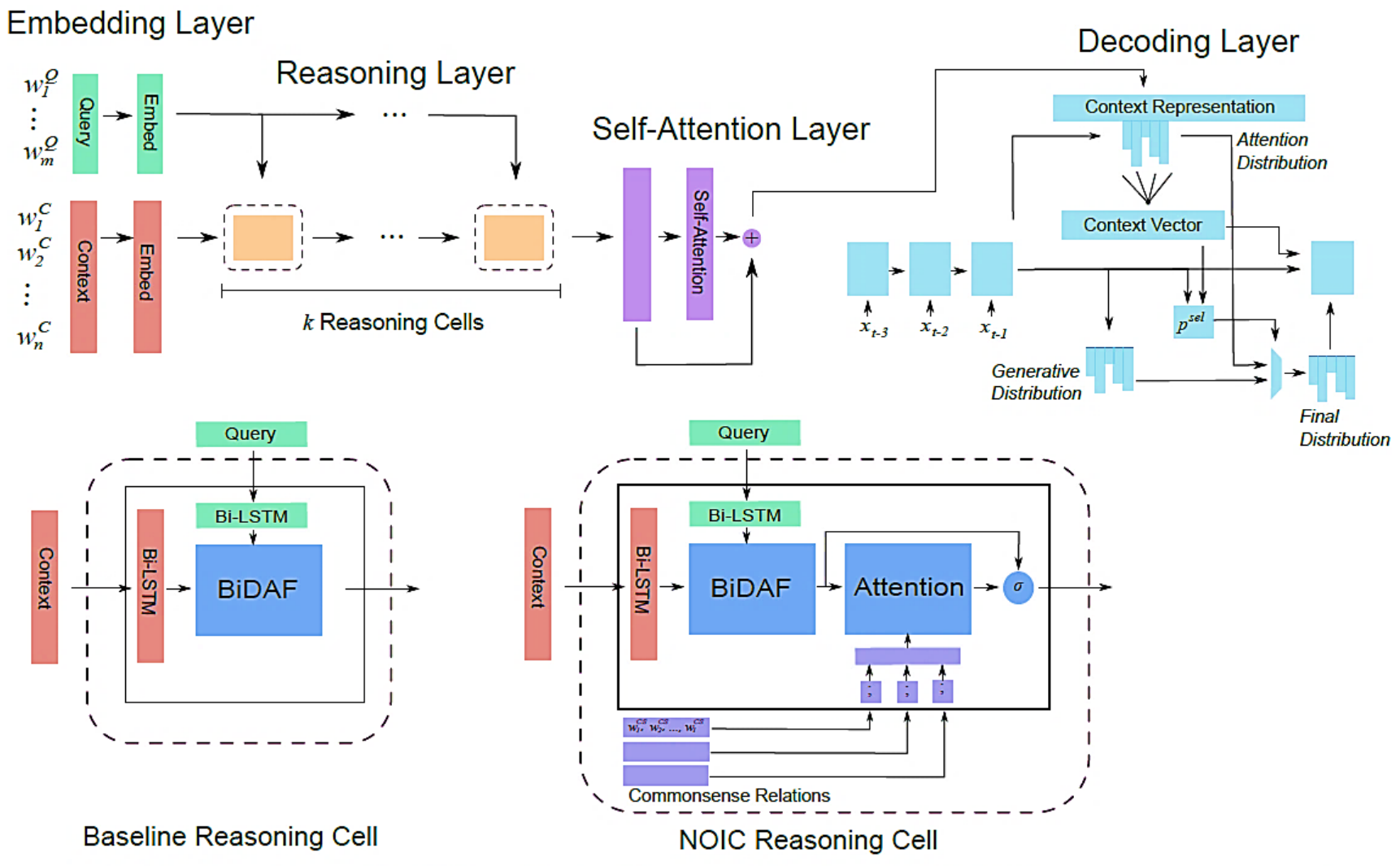

Figure 10: The architecture of MHPGM and the NOIC commonsense reasoning cell [29]

**QFE**: Nishida et al. [30] proposed a model for explainable multi-hop QA named Query Focused Extractor (QFE) which is based on a summarization idea. They use the multi-task learning of the QA model for answer selection and QFE for evidence extraction. QFE as the main part of this model adaptively determines the number of evidence sentences by considering the dependency among the evidence sentences and the coverage of the question. Unlike other approaches that extract each evidence sentence separately, QFE uses an RNN and attention mechanism to consider important information in the question and the relationships between sentences. This query-aware recurrent structure enables QFE to consider the dependency among the evidence sentences and cover the important information in the question sentence. In brief, the main goal of QFE is to summarize the context according to the question. Query-focused summarization is the task of summarizing the source document with regard to the given query. The multi-task learning with QFE is general in the sense that it can be combined with any QA model. The overview of QFE is shown in Figure 11. $z^t$ is the current summarization vector, $g^t$ is the query vector considering the current summarization, $e^t$ is the extracted sentence, $x_e^t$ updates the RNN state.

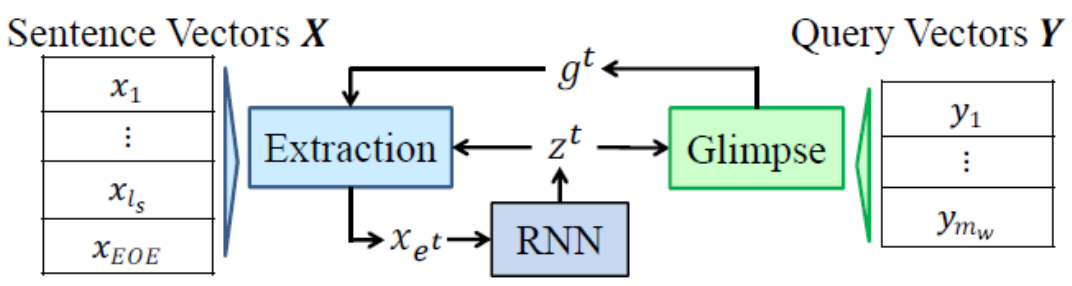

Figure 11: Overview of Query Focused Extractor at step t [30]

**TAP**: Bhargav et al. [31] proposed a deep neural architecture, called TAP (Translucent Answer Prediction) cover of two main ideas: (1) *Local Interaction*: Each sentence should be understood in the context of its neighboring sentences and the question, (2) *Global Interaction*: A global (inter-passage) interaction among sentences must be identified and used for supporting facts. TAP is a hierarchical architecture that tries to capture the local and global interactions between the sentences and consists of two main parts: (Figure 12)

- *Local and Global Interaction eXtractor* (LoGIX) with three layers: local layer to obtain intra-passage dependencies, *Global* Layer to obtain inter-passage dependencies, and *Supporting Facts Prediction* Layer to calculate the probability that a sentence is a supporting fact.
- *Answer Predictor (AP*) to predict the final answer by reasoning over the supporting facts. It consists of four parts: *Input Data Shaping* to preprocess and concatenate the supporting facts, *Context Encoding* to encode the context using a pre-trained BERT model, *Answer Type Predictor* to classify the question into one of the three classes (yes, no and, span), and *Answer Span Predictor* to predict the final answer.

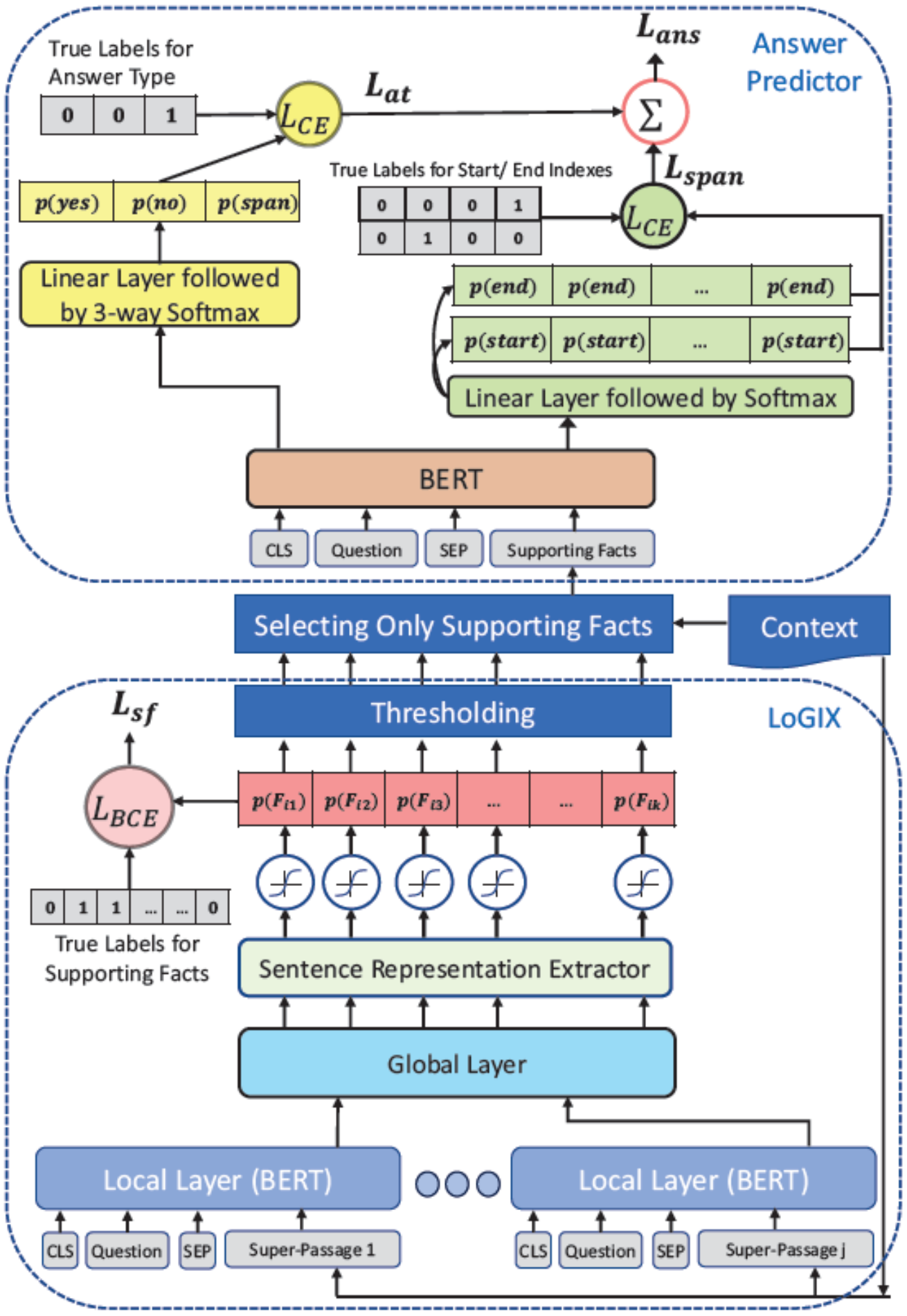

Figure 12: Architecture of TAP [31]

**PH-Model**: Cong et al. [32] focused on using the benefit of the hierarchical structure of the natural language text (document-passage-sentence-word-character), while most existing studies ignore this information in the natural language context. Then they proposed a model for Chinese multi-hop MRC named (PH-Model), in which P stands for the Passage reranking framework, and H denotes the Hierarchical neural network. As you can see in Figure 13, PH-Model consists of multiple main parts: *Bi_ONLSTM*, that is an ordered neuron LSTM is used to obtain hierarchical information from a passage (Instead of traditional LSTM),

Bidirectional Attention Flow is used to extract the hierarchical information form paragraphs to get the query-aware context and the context-aware query representation, *Fused* layer is used to merge all information and finally *Pointer network* to obtain the probability of the start and end positions of the answer

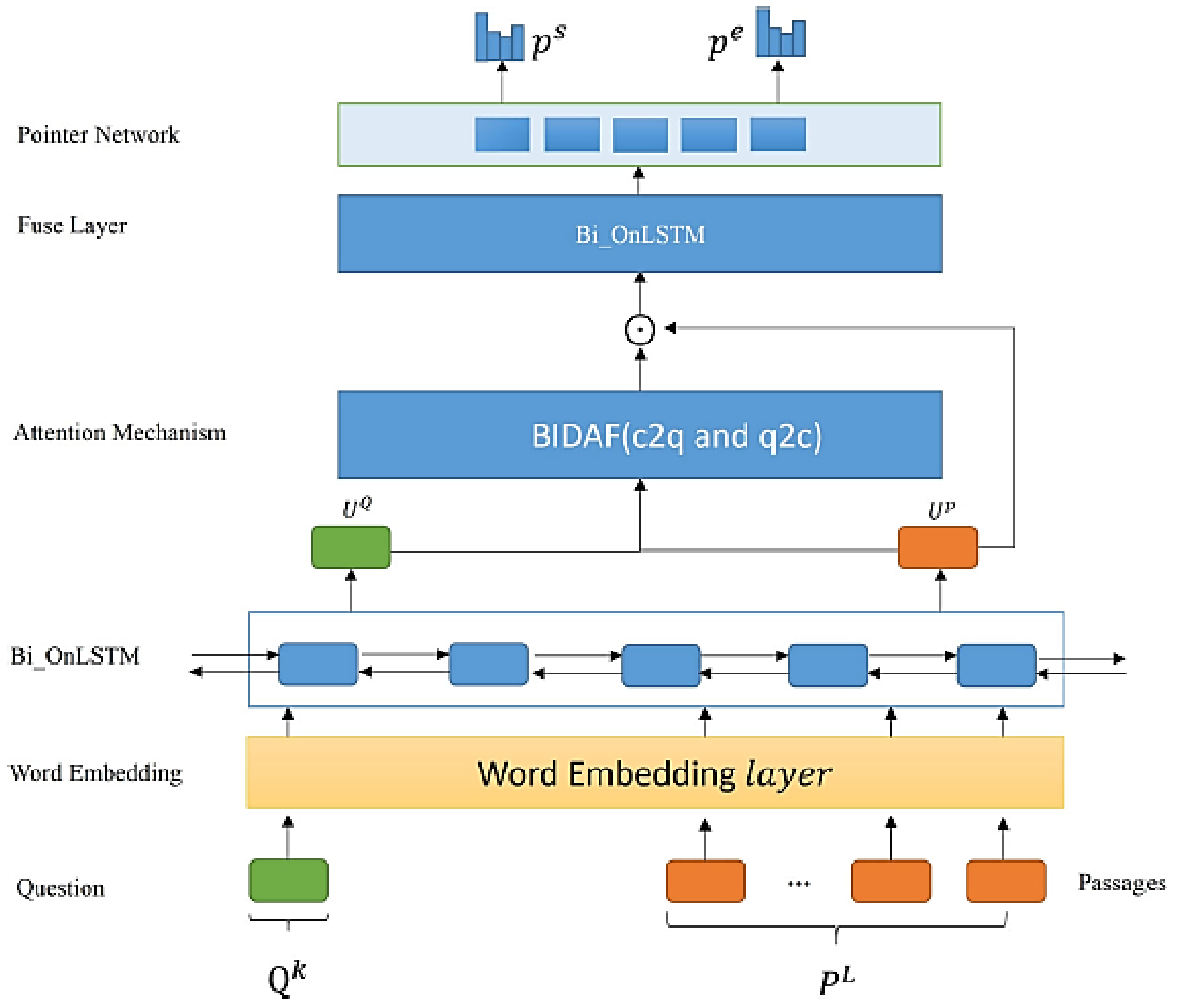

Figure 13: Architecture of PH-Model [32]

### *3.2.1 Path-based:*

In the following, we review models that focus on finding the right path in information to find the answer. As multi-hop MRC faces more information and complex questions, finding the right path has become more important and difficult, so many models in this technique are path-based. One of the important advantages of path-based models is that they are interpretable because they can provide the evidence chain to the final answer.

**EPAr**: Jiang et al. [33] proposed an interpretable model named Explore-Propose-Assemble reader (EPAr) to mimic the coarse-to-fine-grained reasoning behavior of humans when facing multiple long documents. The main idea is to construct a reasoning tree according to the documents like a hierarchical memory network and use the path in this tree to extract the final answer. This model has three components as shown in Figure 14:

- The *T-hop Document Explorer (DE)* module constructs the reasoning tree like a hierarchical memory network. In each step, it selects one document, updates the memory cell using the selected document, and iteratively selects the next related document.
- The *Answer Proposer (AP)* module uses the constructed reasoning tree to predict an answer from every root-to-leaf path.
- The *Evidence Assembler (EA)* module extracts a key sentence containing the proposed answer from every path and combines them to predict the final answer.

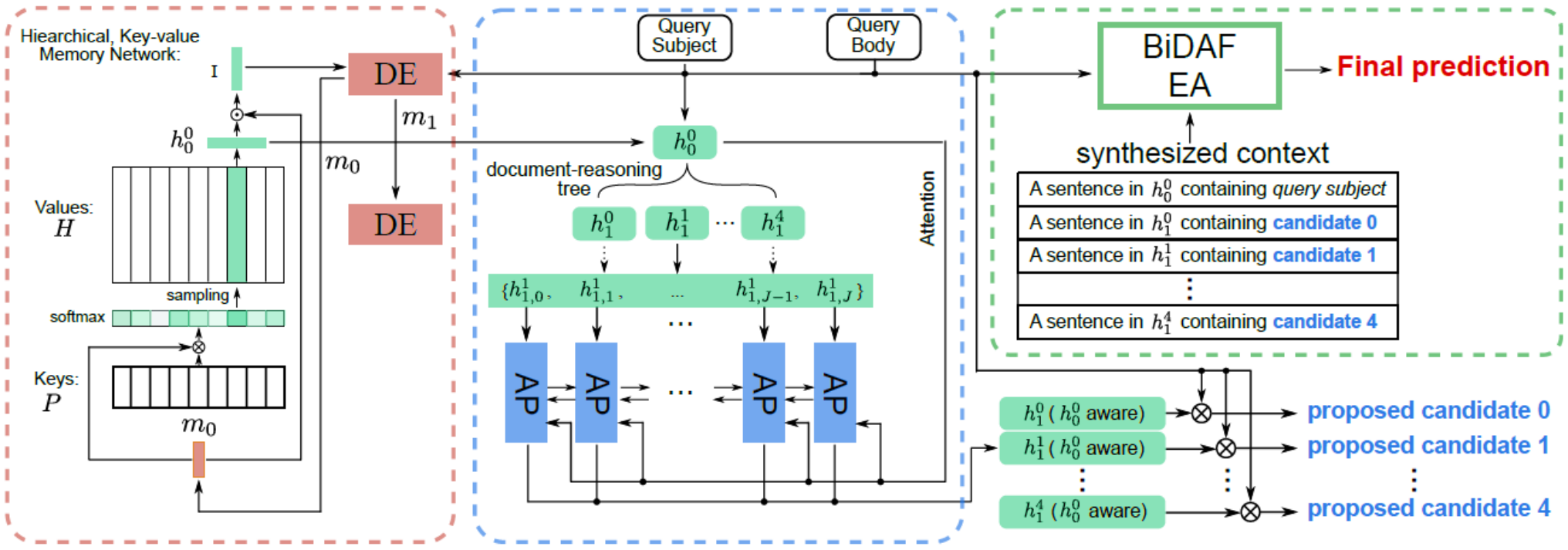

Figure 14: Architecture of EPAr [33]

**PathNet**: Kundu et al. [34] proposed a path-based reasoning approach for multi-hop MRC which first extracts all paths in the passages based on implicit relations between entities, and then composes the passage representations along each path to compute a passage-based representation. In other words, the passages representation is achieved by considering the paths.

They first find all possible path from passages. It starts with selecting a passage that contains a head entity from the question, and then finds all entities and noun phrases from the same sentence. Afterward, it selects the next passage that contains the potential intermediate entity identified above. Finally, it is checked whether the next passage contains any of the candidate answer choices or not. The resulting will be a set of entity sequences. After obtaining all potential paths, it is time to score each path using the *PathNet* model based on two perspectives: 1) *Context-based Path Scoring*, which is based on the interaction with the question encoding, and 2) *Passage-based Path Scoring*, which is based on the interaction between the passage-based path encoding vector and the candidate encoding. There is an example of the process in Figure 15 which In the Rank-1 path, the model composes the implicit located in relations between (Zoo lake, Johannesburg) and (Johannesburg, Gauteng). However, this method extracts many invalid paths, then causes wasting the computing resources [35].

| |
|---|
| **Question**: (zoo lake, located in the administrative territorial entity, ?)<br>**Answer**: gauteng |
| **Rank-1 Path**: (zoo lake, Johannesburg, gauteng)<br>Passage1: ... **Zoo Lake** is a popular lake and public park in **Johannesburg**, South Africa. It is part of the Hermann Eckstein Park and is ...<br>Passage2: ... **Johannesburg** (also known as Jozi, Joburg and Egoli) is the largest city in South Africa and is one of the 50 largest urban areas in the world. It is the provincial capital of **Gauteng**, which is ... |
| **Rank-2 Path**: (zoo lake, South Africa, gauteng)<br>Passage1: ... **Zoo Lake** is a popular lake and public park in Johannesburg, **South Africa**. It is ...<br>Passage2: ... aka The Reef, is a 56-kilometre - long north - facing scarp in the **Gauteng** Province of South Africa. It consists of a ... |

Figure 15: Two top-scoring path for an example

**DRL-GRC**: Long et al. [36] focused on the inherent sequential of the multi-hop MRC, which means the system must decide where to look next, based on the current state. Then they proposed a model named Deep Reinforcement Learning-based where the knowledge extraction phase is explicitly decoupled from the question answering phase. This model uses the current information in the knowledge chain to inform which information should be achieved in the next step. The model consists of two main components:1) a *Linker* to construct a sentence-level chain form the sentences of supporting documents that allow movement between documents, and 2) an *Extractor* which learns where to look based on the current question and already-visited sentences.

As Figure 16 shows, after embedding sentences into a matrix and constructing a graph, the policy is run to select the next sentence. In this regard, the convolution-based policy network is used to learn a traversal policy from current options, previously accepted sentences, and the current question. For example, in this figure, the policy has learned to select the correct next sentence ($c_t$) from the previous sentence ($c_{t-1}$). After this action, the answer sentence ($o_1$) can be selected during the next step.

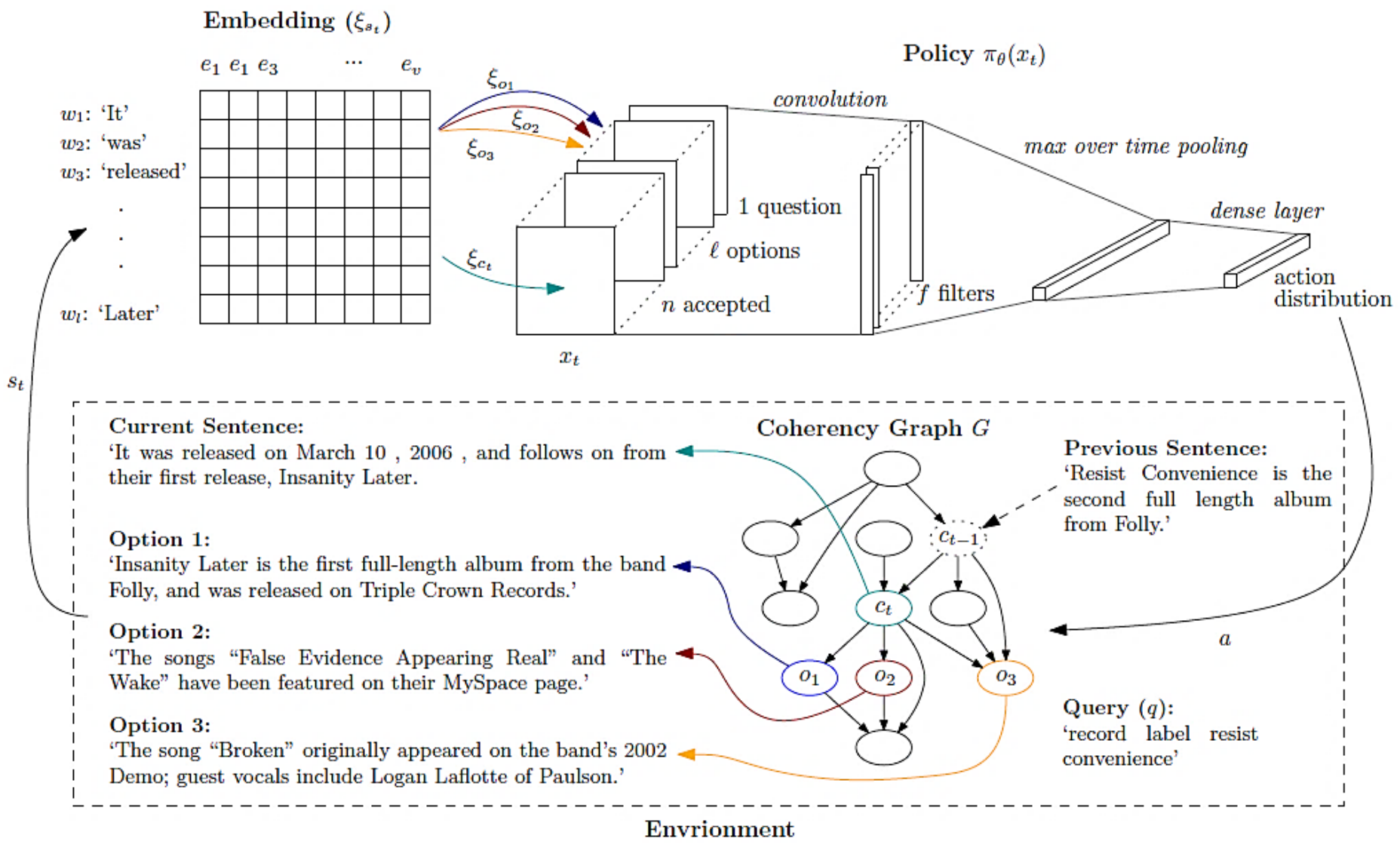

Figure 16: Architecture of DRL-GRC [36]

**SMR**: Huo, Ge and Zhao [37] proposed a Sentence-level Multi-hop Reasoning approach named SMR while most existing approaches only use document-level or entity-level inferences. In this regard, an initial sentence is first found based on the main entity in the question, and that sentence is then used to find more relevant sentences to create multiple sentence-based reasoning chains as a memory network. Besides, some sentences are concatenated to prevent the mistakes of Co-Reference resolution methods and to reduce the number of hops. An example of such a concatenation is shown in Figure 17.

There are two phases in this model: 1) the *Selecting* phase to select the most relevant sentence to the network memory state as the initial sentence of the current hop, and 2) the *Establishing* phase to prepare to go to the next hop by updating the network memory state. Finally, the information of the reasoning chains is used to predict the final answer.

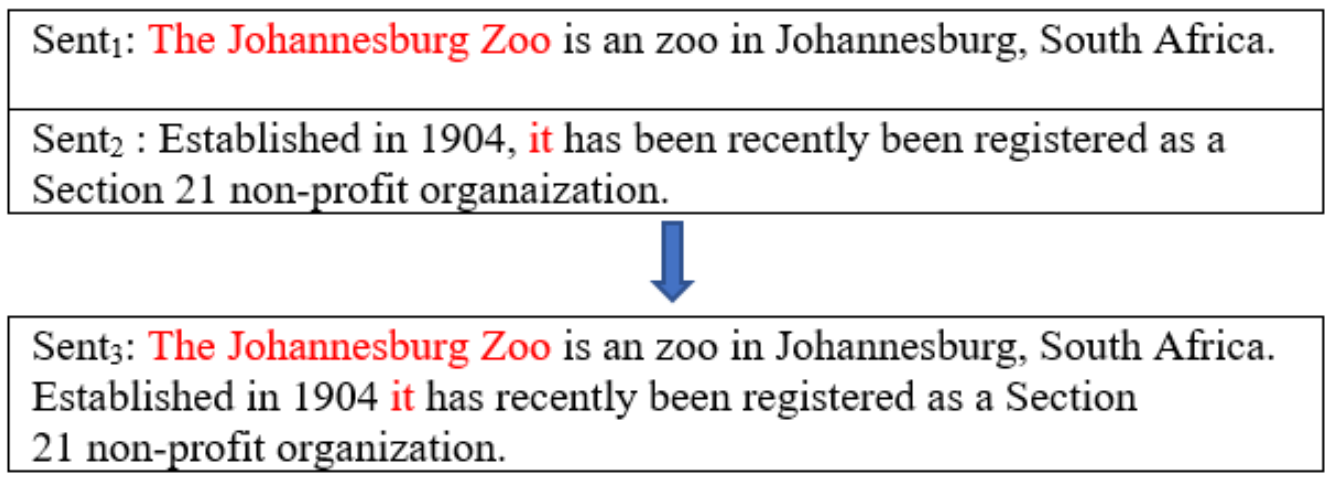

Figure 17: Representation enrichment via concatenation of adjacent sentences [37]

**ChainEx**: Chen, Lin and Durrett [38] proposed a sentence-based model that does not rely on gold annotated chains or supporting facts at the training and test phases, instead, pseudo-gold reasoning chains are derived using some heuristics based on named entity recognition and coreference resolution during the training time, and it learns to extract chains from raw texts at the test time. They first extract a reasoning chain over the text for a multi-hop reasoning task, and then apply a BERT-based QA system to find the answer by learning from these chains.

To construct the reasoning chain, each sentence is considered as a node in the chain, and there is an edge between two sentences if they have the same entity. Besides, there are edges between all sentences from the same paragraph. The model starts from the question and finds all possible reasoning chains. The chain extractor is a neural network with two main components: *Sentence Encoding* and *Chain Prediction*. In the sentence encoding component, the BERT-Para model provides a representation from each paragraph jointly with the question. In the chain prediction component, an LSTM-based pointer network is used to extract the reasoning chain. The output of the chain extractor is a variable-length sequence of sentences. Finally, a BERT-based QA system is applied to the extracted chains to find the final answer.

Figure 18 show an example with two possible “reasoning chains”. The first chain is most appropriate, while the second requires a less well-supported inferential leap.

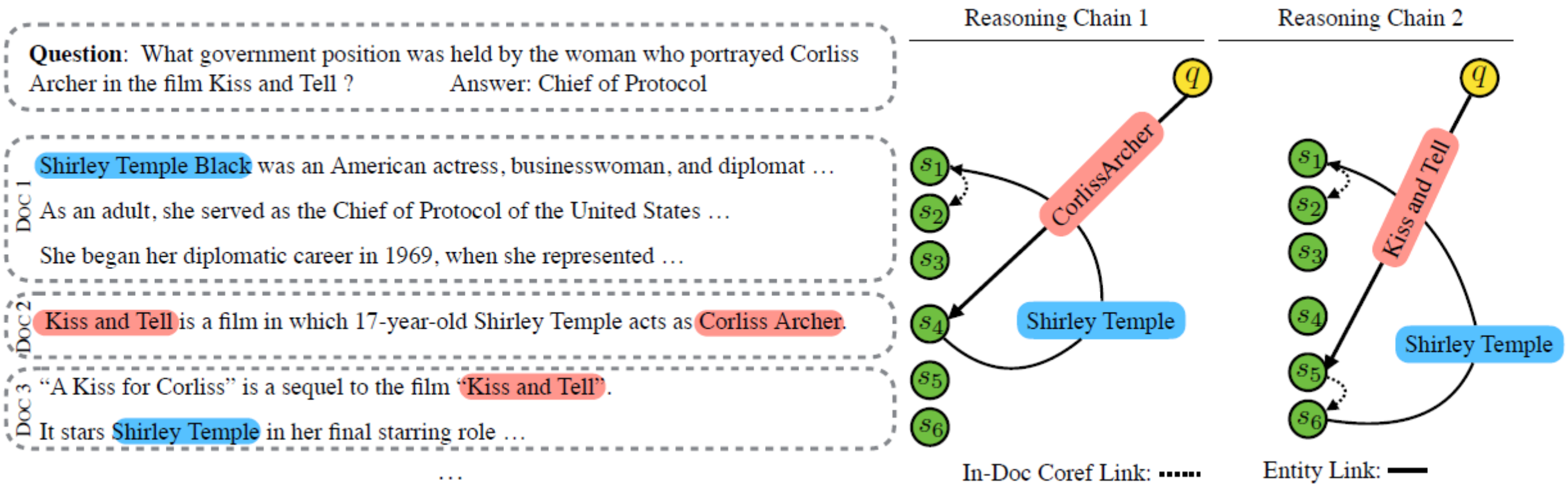

Figure 18: A multi-hop example in ChainEx [38]

**SCR**: Huo and Zhao [35] proposed another sentence-level model, since using entity and document makes the chain too coarse and ambiguous or too subtle, limited, and less accurate. This study believes that using sentences as inference nodes is more reasonable than using documents or entities. To explain the problem, you can see an example in Figure 19, as it is clear, with using documents, although the path is more complete, there is a lot of unrelated information, which can produce redundancy. On the other hand, with using entities, although the path is concise it is too subtle and limited, and much information will be lost. However, if the path uses sentences as nodes, it is not limited anymore and with less information redundancy. Besides a sentence-based path can explain the reasoning process better.

Then they proposed a Sentence-based Circular Reasoning (SCR) approach, and it consists of three modules*: Sentence Encoder (SE)* to obtain the sentence representation, *Path Generator (PG)* to iteratively infer among sentences of multiple documents according to the question, and *Path Evaluator (PE)* to evaluate the obtained path and predicts an answer (Figure 20).

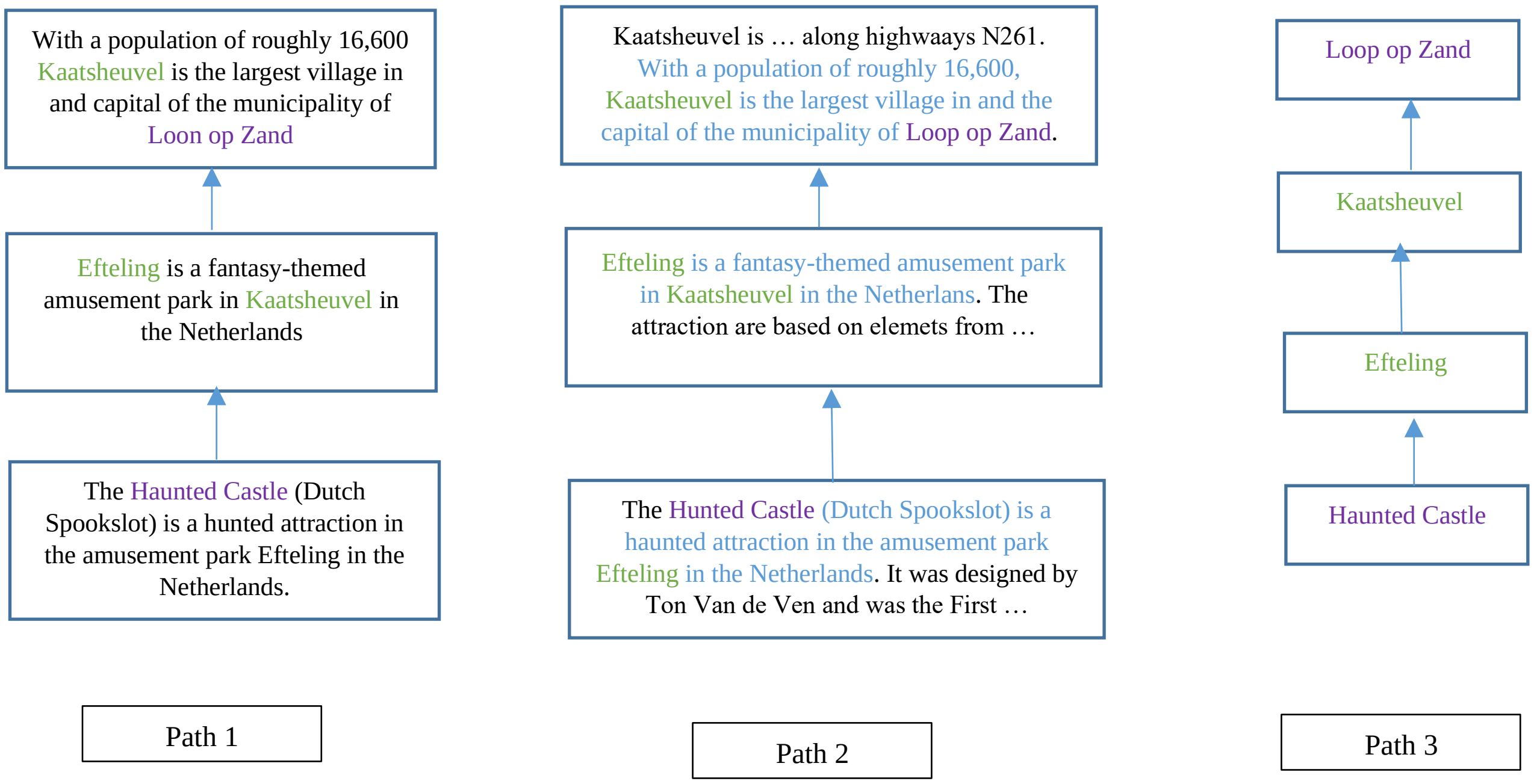

Figure 19: Paths of different granularities [35]

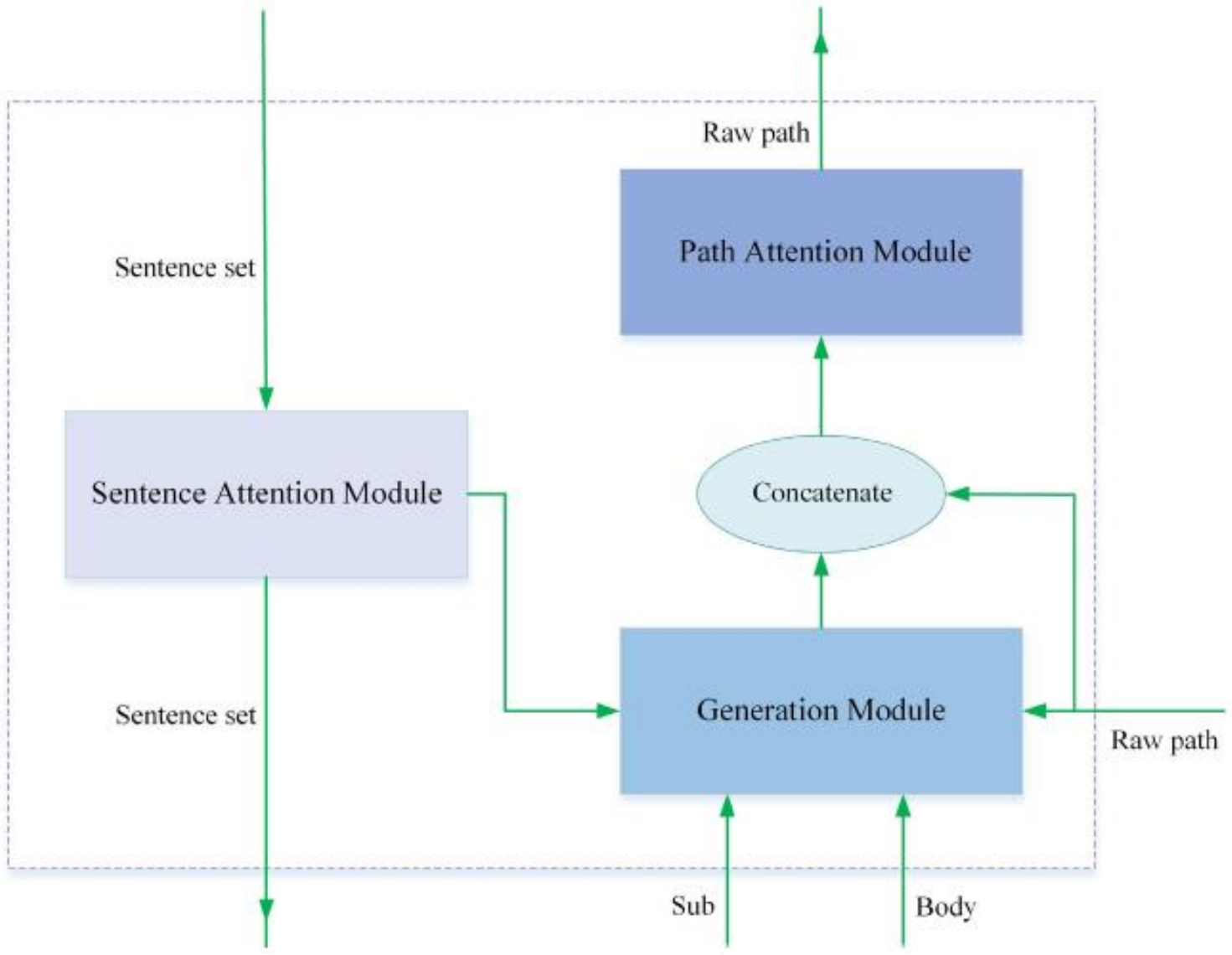

Figure 20: Architecture of SCR [35]

Unlike the decomposition technique, this technique focuses on multi-hop reasoning instead of simplifying the multi-hop MRC problem and has achieved great attention among studies. The main disadvantage of this technique is that they are expensive and time-consuming approaches, and require a large amount of data to train.

### 3.3 Graph-based technique

Graph-based techniques because of their natural language relationship representation ability [39] has attracted attention in multi-hop MRC. It is natural to model natural language context into graph structure and the process of multi-hop reasoning as moving among nodes [14]. The main idea of the Graph-based technique is to construct a graph based on the context and question, and then the reasoning is performed by message passing over this structure using graph neural networks. The process of constructing the graph from large textual data and reasoning over it are challenging tasks. There are a lot of studies that focus on these challenges which are explained in this subsection in detail.

Constructing graphs from input data is one of the basic parts of this technique. Some studies construct an entity graph from the input data, which means the nodes of the graph nodes are the entities of the context and question. A lot of studies work on this kind of graph which will be reviewed in the following.

#### *3.3.1 Entity-node graphs:*

One of the most basic ideas for converting text to a graph is to construct an entity graph, which means the entities within the text are node graphs and the relationships between the entities are considered edges. In this section, the studies that uses entity graph will be reviewed.

**MHQA-GRN**: Song et al. [40] focused on inferring global context as an important key in multi-hop reading comprehension, while previous studies approximate global evidence with local coreference information with DAG-styled GRU. They proposed a model for better connecting global evidence, with a more complex graph compared to DAGs. They construct an entity graph with three types of edges: the edge between the same entity within a passage, the edges between two mentions of different entities within a context window, and coreference-typed edges. The graph might also have cycles which makes it difficult to apply a DAG network to it. (A graph with three types of edges and a DAG graph are shown in Figure 21).

For inferring the global context, the related information of the constructed graph has been merged. In this study, two recent graph neural networks have been applied to this graph: graph convolutional network (GCN) and graph recurrent network (GRN) for evidence aggregation. Afterward, an attention mechanism is applied in order to match the hidden states at each graph encoding step with the question representation. Finally, a probability distribution is calculated from the matching results. The architecture of this model is shown in Figure 22. However, this model still only implicitly combines knowledge from all passages, and are therefore unable to provide explicit reasoning paths [28].

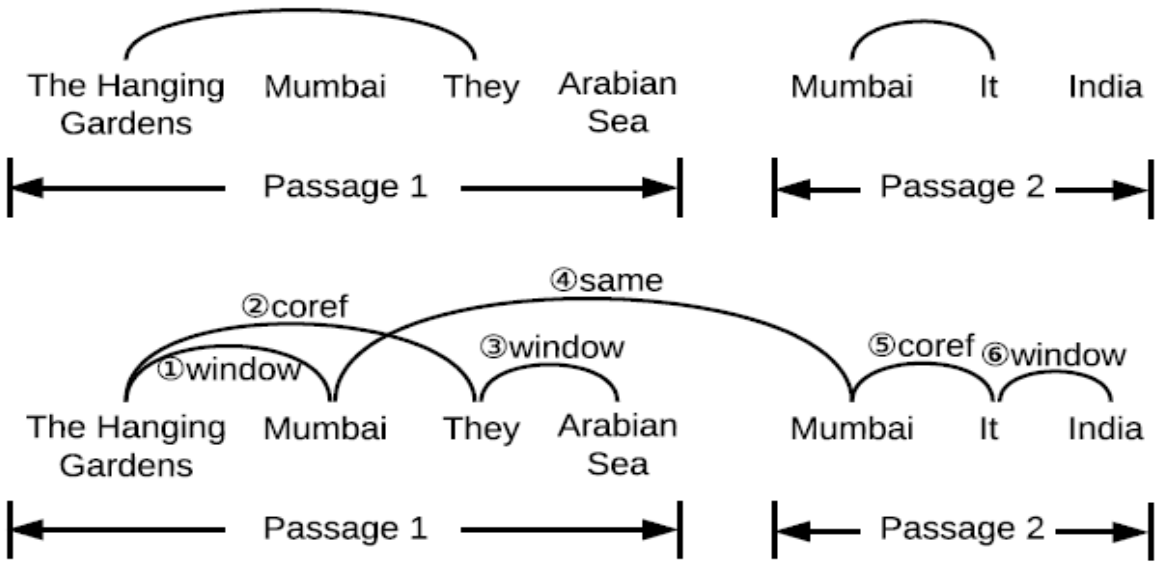

Figure 21: A DAG graph (top) and a graph by considering all three types of edges (bottom) [40]

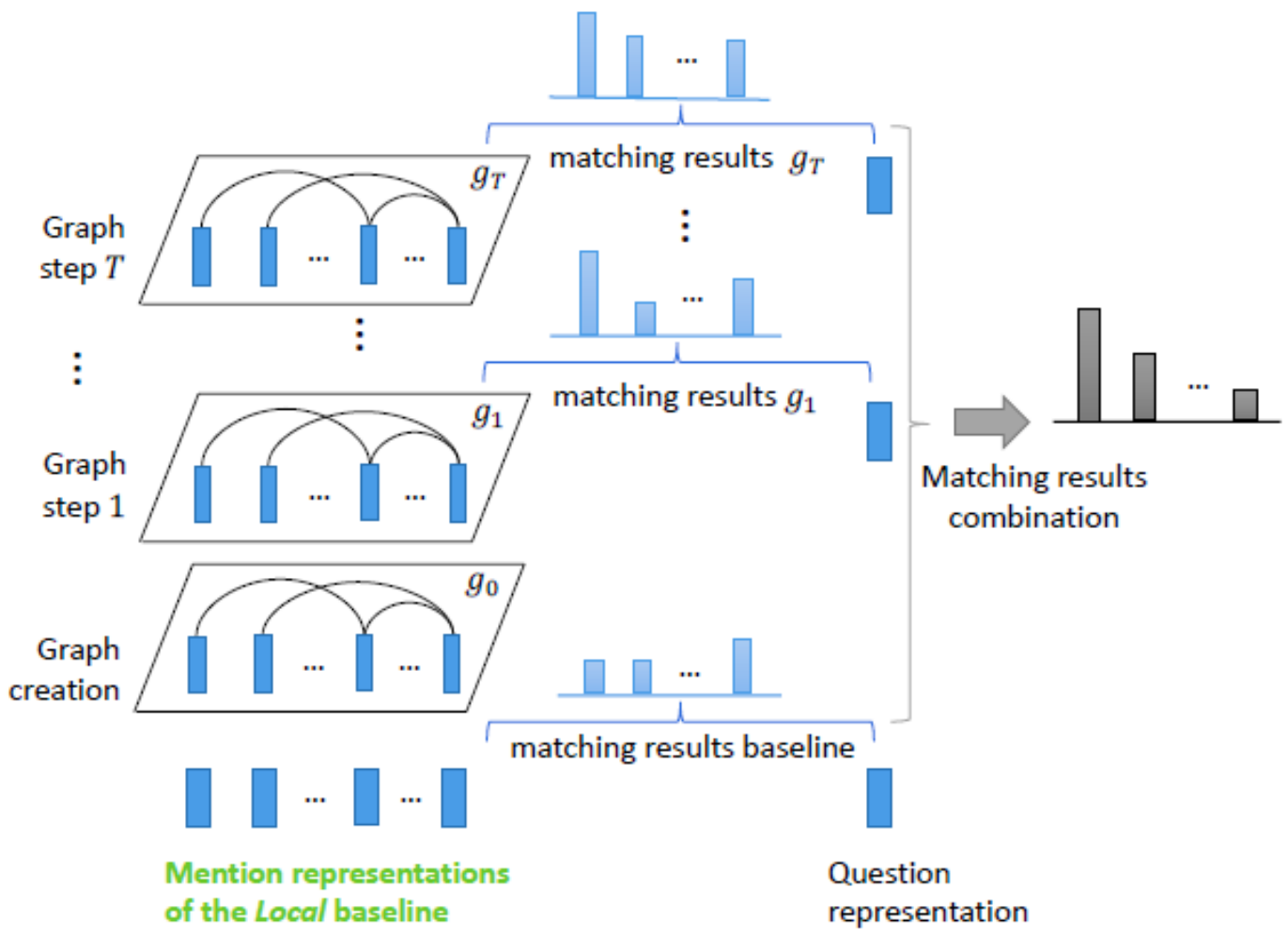

Figure 22: Architecture of MHQA-GRN [40]

**DFGN**: Xiao et al. [41] proposed a model to improve the interaction between the information of documents and the entity graph. They proposed a fusion process of Doc2Graph and Graph2Doc for multi-hop reasoning that leads to a less noisy entity graph and more accurate answers. The process of constructing dynamic entity graph iterates in multiple rounds to achieve multi-hop reasoning. In each round, DFGN generates and reasons on a dynamic graph, where irrelevant entities are masked out while only reasoning sources are preserved, via a mask prediction module. Then the fusion block not only aggregates information from documents to the entity graph (doc2graph) but also propagate the information of the entity graph back to document representations (graph2doc). Figure 23 illustrates the *Fusion block* in DFGN which consists of:

- Passing information from tokens to entities by computing entity embeddings from tokens (Doc2Graph flow).
- Propagating information over the entity graph.
- Passing information from the entity graph to document tokens, as the final prediction is on tokens (Graph2Doc flow).

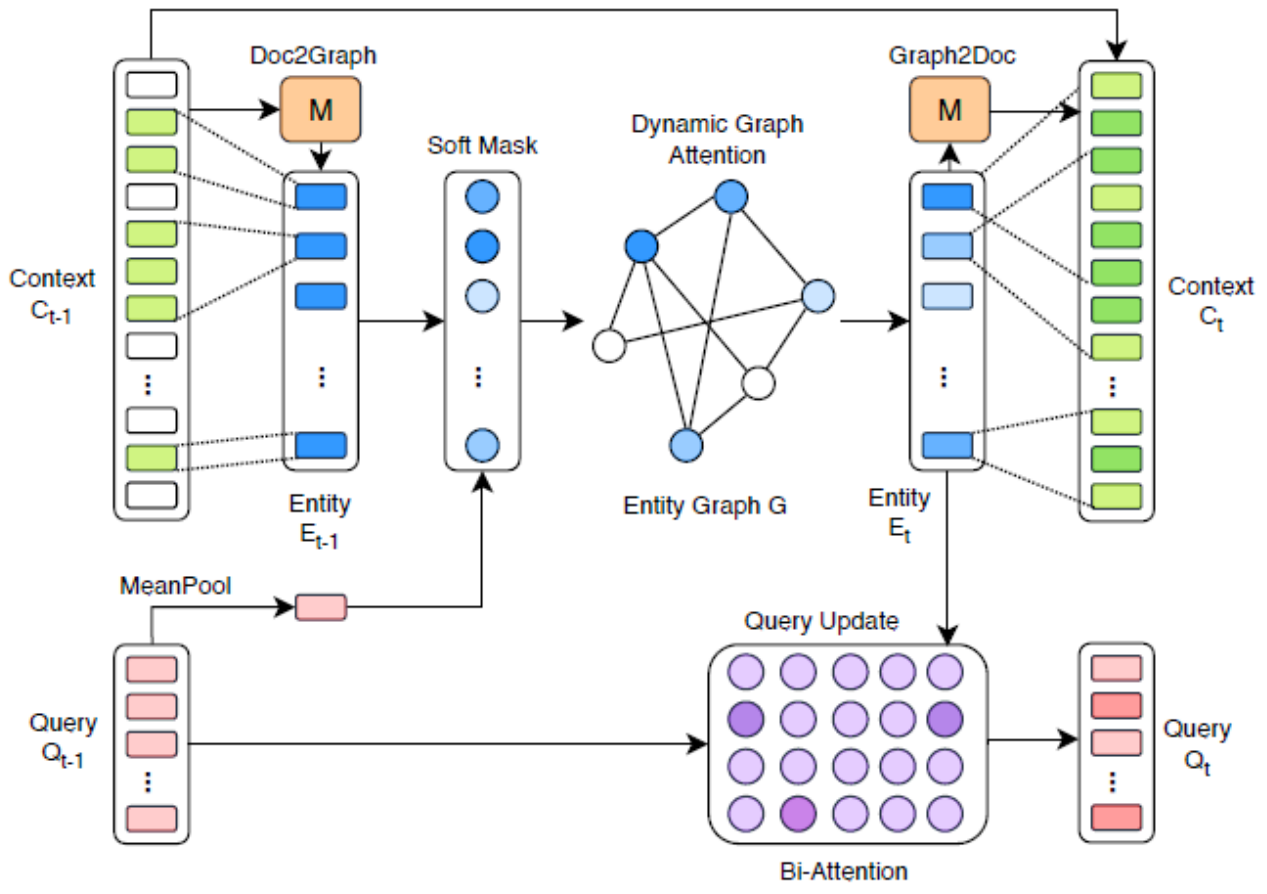

Figure 23: Architecture of the Fusion block [41]

**Entity-GCN:** De Cao, Aziz, and Titov [42] proposed a method named Entity-GCN to provide an efficient training technique that relies on a slower offline and a faster online computation that does not require expensive document encoding. Since in multi-hop MRC, models often face multiple passages (e.g., Wikipedia or a domain-specific set of documents) using expensive document encoders like RNN or transformer-like self-attention won't be efficient. In this approach, only a small query encoder, the GCN layers, and a simple feed-forward answer selection component are learned. Also, instead of training RNN encoders, a contextualized embedding (ELMo) is used to obtain initial (local) representations of nodes, as a result, only a lightweight computation has to be performed online and the rest is can be done offline. The result shows that in the WikiHop [17] dataset, the model is at least 5 times faster to train than BiDAF. Figure 24 shows an example of the constructed graph in the Entity-GCN model. In this example, there are three documents, each of which is indicated by a graph within dashed ellipses. The nodes in this graph are entities. Nodes with the same color refer to the same entity, solid edges are co-occurrence edges in the same document, dashed edges are those mentions that are exactly matched and finally, bold-red lines are coreferences.

However, this model still only implicitly combines knowledge from all passages, and are therefore unable to provide explicit reasoning paths [28].

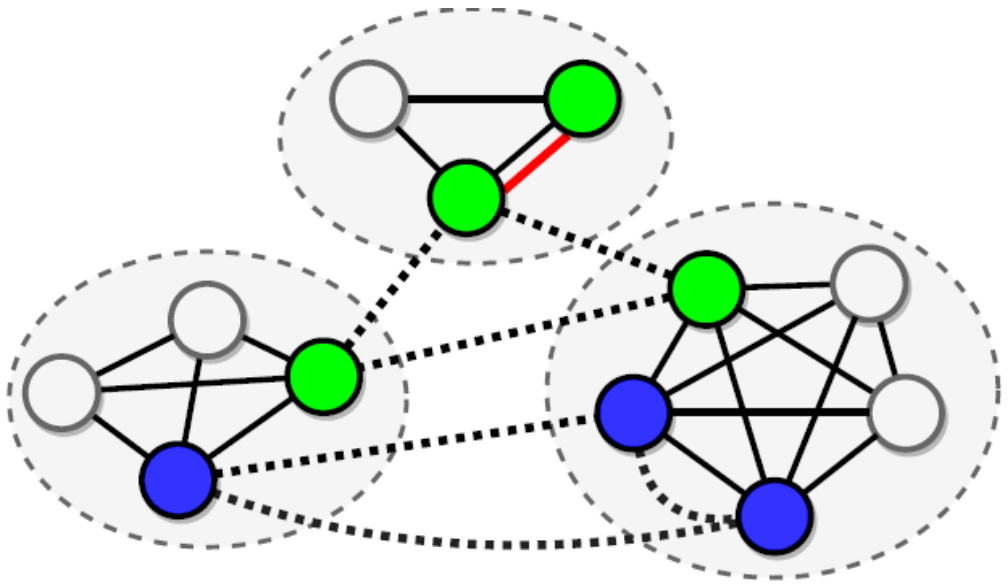

Figure 24: An example of the constructed graph in Entity-GCN [42]

**BAG**: Cao, Fang, and Tao [43] proposed a Bi-directional Attention Entity Graph Convolutional Network (BAG) with a focus on Relational Graph Convolutional Network (RGCN) to realize multi-hop reasoning by message propagating across different entity nodes in graphs and generating transformed representations of original ones. The R-GCN is employed to handle high-relational data characteristics and make use of different edge types. It uses graph neural networks to obtain the relationship between entities, or add a self-attention mechanism into the model, so as to obtain a gain in the result. They first construct an entity graph and apply graph convolutional networks to obtain a relation-aware representation of nodes. Then, the bidirectional attention mechanism is used to generate the representation of query-aware nodes. As it is shown in Figure 25, the model has five layers, *Entity Graph Construction*, *Multi-level Features*, *GCN*, *Bi-directional Attention*, and *Output*. However, as the number of inferences increases, the complexity of models will rise sharply due to the iteration of the message passing algorithm, resulting in low efficiency [35].

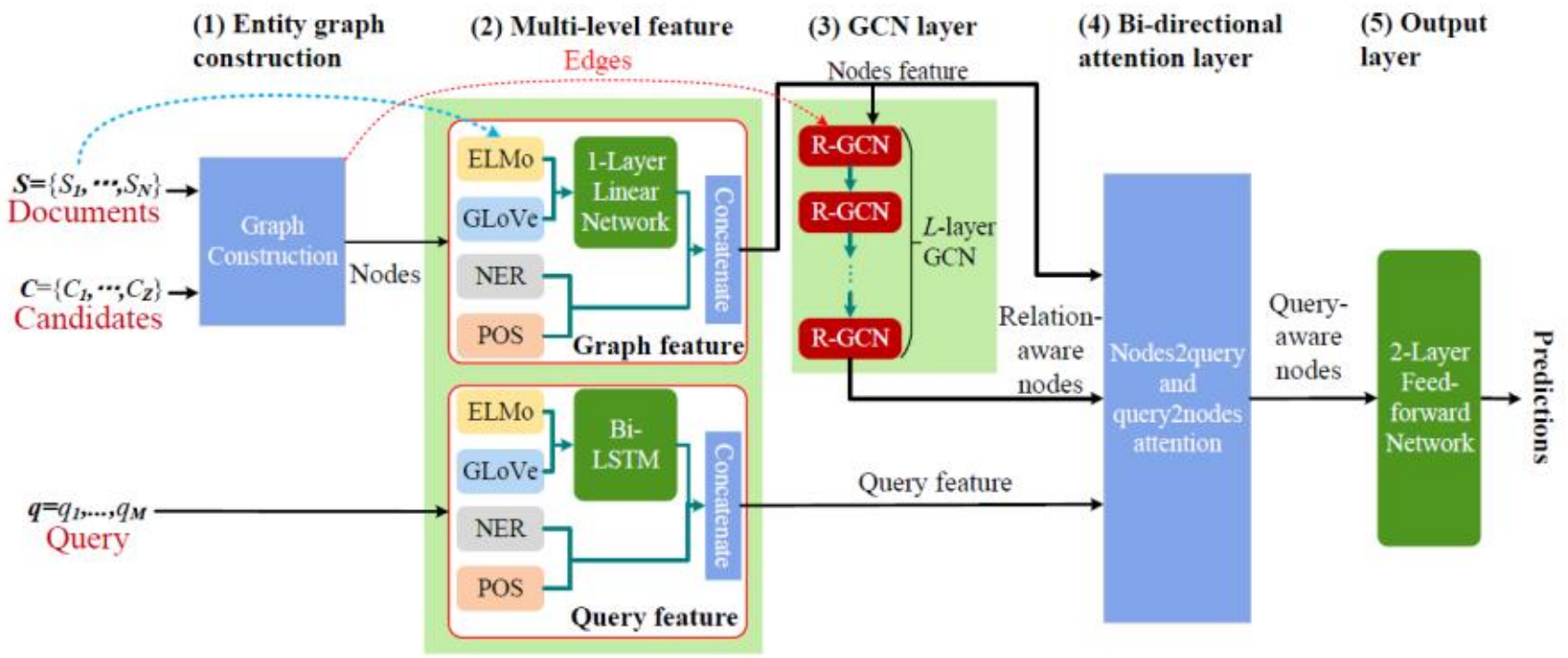

Figure 25: Architecture of BAG [43]

CogQA: Ding et al. [44] to mimic the process of the human brain, proposed the CogQA framework that builds a cognitive graph inspired by dual-process theory. This theory approves that the human brain first retrieves relevant information implicitly and unconsciously, and finally an explicit and conscious reasoning process is applied to that relevant information. The cognitive graph structure in this framework can offer ordered and entity-level explainability and suits relational reasoning. Based on this theory, the proposed model has two components:

- *Implicit Extraction (System 1)* in which the relevant information, like question-relevant entities, and candidates answer, are extracted from paragraphs. Then, the extracted relevant entities are used to construct a cognitive graph. Such a graph can be seen as working memory.
- *Explicit Reasoning (System 2)* in which the reasoning procedure is applied to the graph to guide System 1 to extract the next-hop entities. The main part of System 1 is BERT, and the main part of System 2 is GNN.

As it is shown in Figure 26, System 1 generates new hop and answer nodes based on clues[x; G] discovered by System 2, where G is the cognitive graph, and clues[x; G] comprises sentences in paragraphs of x's predecessor nodes from which x is extracted.

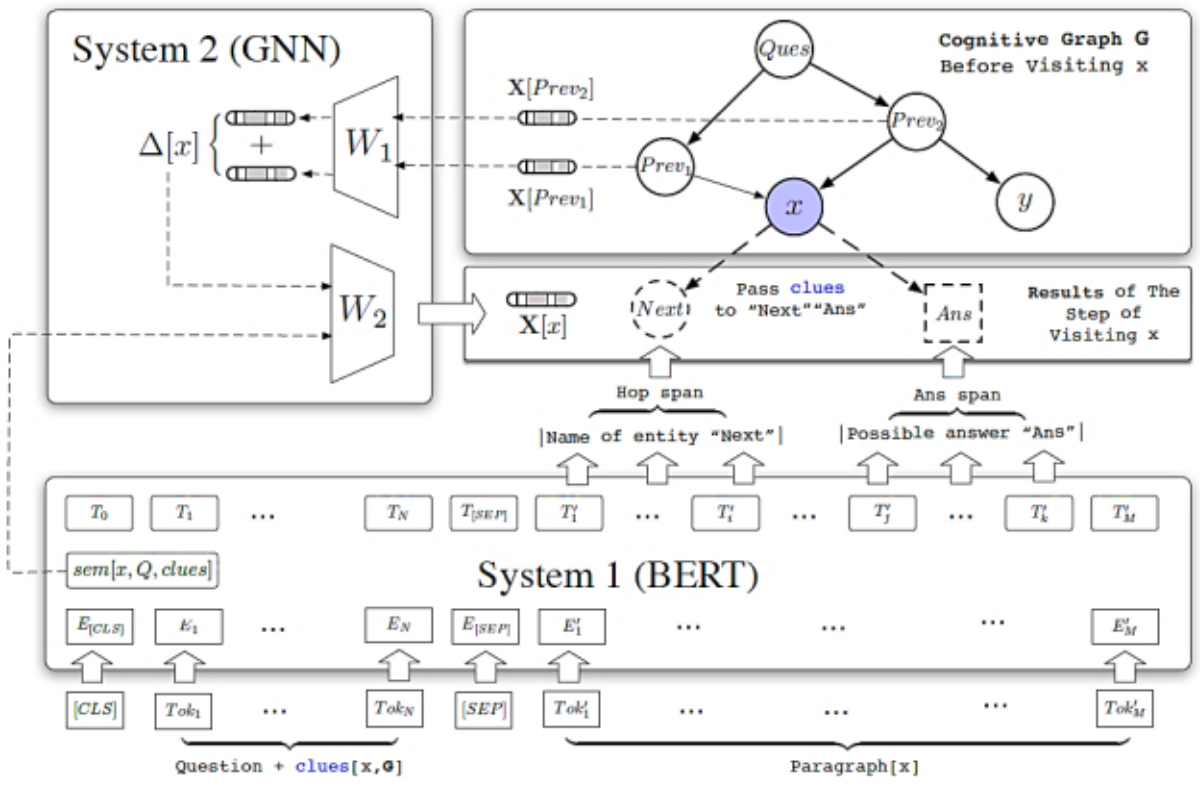

Figure 26: Architecture of CogQA [44]

DRNQA: Li et al. [45] proposed a Dynamic Reasoning Network (DRN) approach that, unlike other models that read the question only once, uses a query reshaping mechanism that considers the question several times, which as a result enhances the ability of the reasoning over the entity graph. There are two main parts:

The entity graph construction component to construct the entity graph (Figure 27). This graph is constructed from three levels: 1) Question-based level: there is an edge between two entities if their sentences have a common entity or phrase with the question, 2) Context-based level: there is an edge between two entities if they are from the same passages, and 3) Passage-based level: there is an edge between two entities if their sentences have at least one entity or phrase in common.

The dynamic reasoning network component (Figure 28) to reason over the entity graph. The query reshaping mechanism (Figure 29) causes the important parts of the question to be read frequently. In this mechanism, a weight is given to different parts of the question repeatedly. In this regard, a Graph Neural Network (GNN) and a Dynamic Graph Attention (GAT) are used in this component. As you can see in Figure 64.

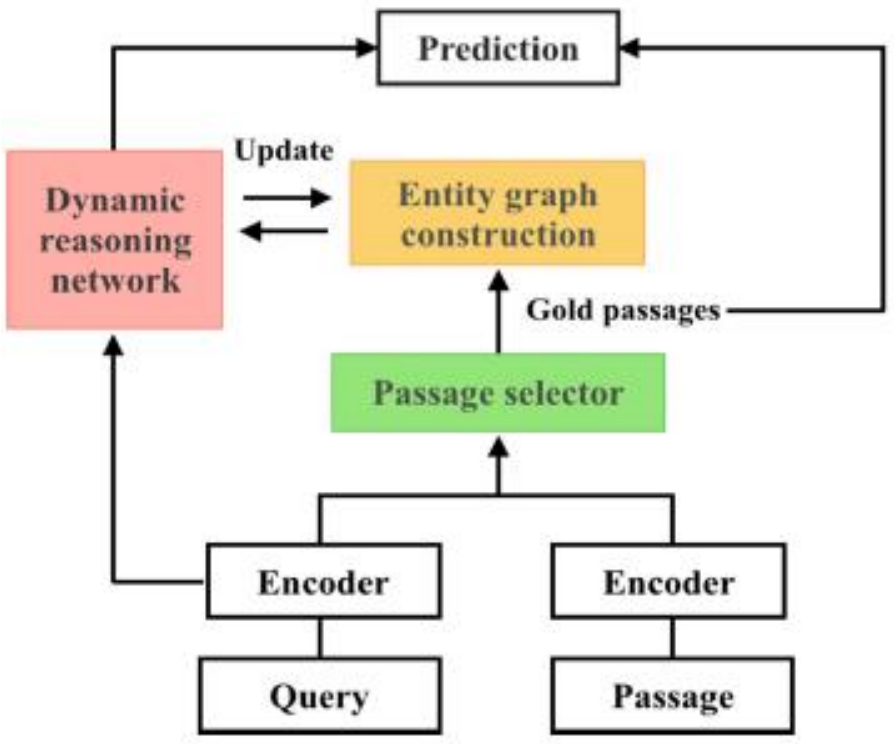

Figure 27: Architecture of DRNQA [45]

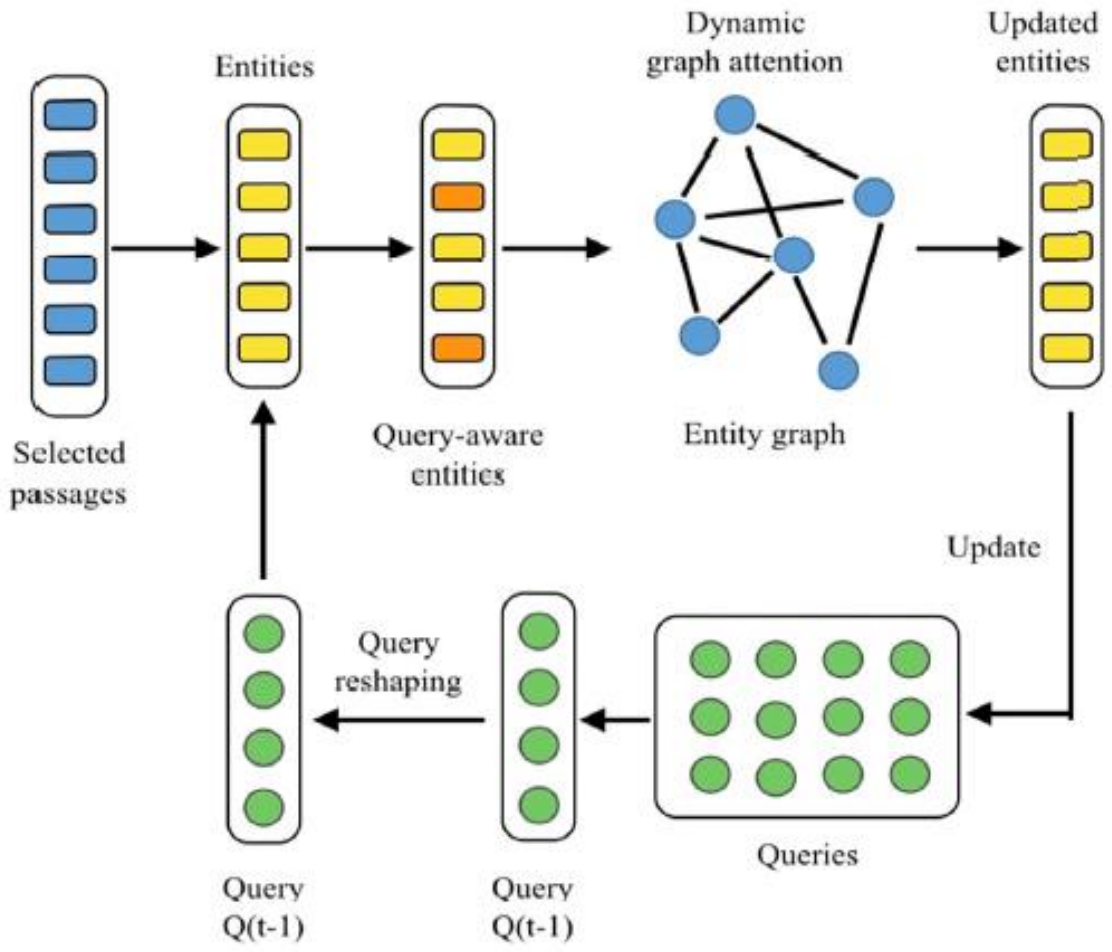

Figure 28: Dynamic Reasoning Network (DRN) [45]

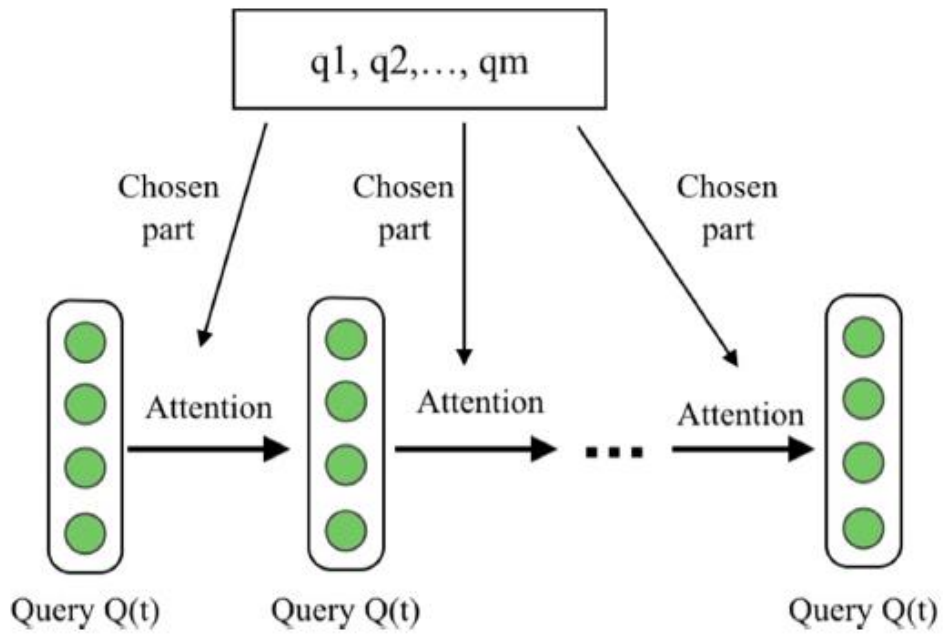

Figure 29: Query Reshaping [45]

### *3.3.2 Multiple-node graphs:*

Previous studies have used entity graphs in which only entities are considered node graphs. However, due to the complex and different structures in natural language text, it seems that entity graphs can cause a lot of information loss. Due to the importance of this issue, many studies have tried to construct more complex and enriched graphs to capture more information about the context, which will be investigated in the following.

**HDE**: Tu et al. [46] proposed a more complex graph named Heterogeneous Document-Entity (HDE) graph with different types of nodes and edges. This graph can cover different granularity levels of information in context and also enables rich information interaction among different types of nodes for accurate reasoning. The nodes in the HDE graph are candidates, documents, and entities. Besides, it has seven types of edges: 1) between a document node and a candidate node that appears in the same document. 2) between a document node and its entity node. 3) between a candidate node and its entity node. 4) between two entity nodes from the same document. 5) between two entity nodes from different documents but they are mentions of the same candidate or query subject. 6) All candidate nodes connect with each other. 7) Entity nodes that do not meet previous conditions are connected as well.

Figure 30 is an example of an HDE graph. In this figure, green nodes indicate documents, yellow nodes denote candidates, and blue nodes stand for entities. In addition, dash lines indicate type 1 edges, dash-dotted lines denote type 2 edges, square dot lines indicate type 3 edges, the red line denotes type 4 edge, the purple line indicates type 5 edge, and black lines indicate type 6 edges. The type 7 edge is not shown in this figure. As Figure 31 shows This model can be categorized into three parts: initializing HDE graph nodes with co-attention and self-attention-based context encoding, and reasoning over HDE graph with GNN-based message passing algorithms and score accumulation from updated HDE graph nodes representations.

However, [35] have shown that if the number of inferences increases, the complexity of models will rise sharply due to the iteration of cumbersome message passing algorithm, resulting in low efficiency.

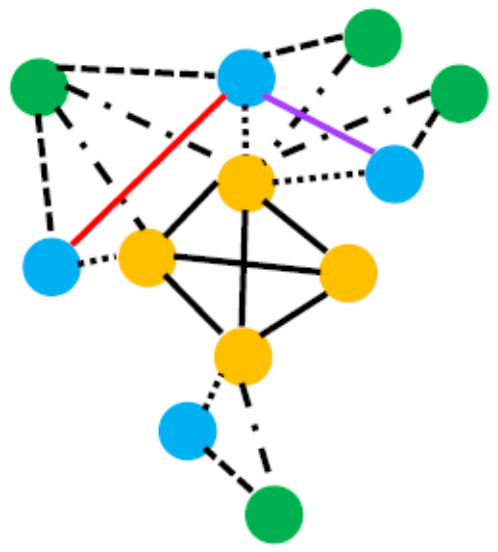

Figure 30: An example of the HDE graph [46]

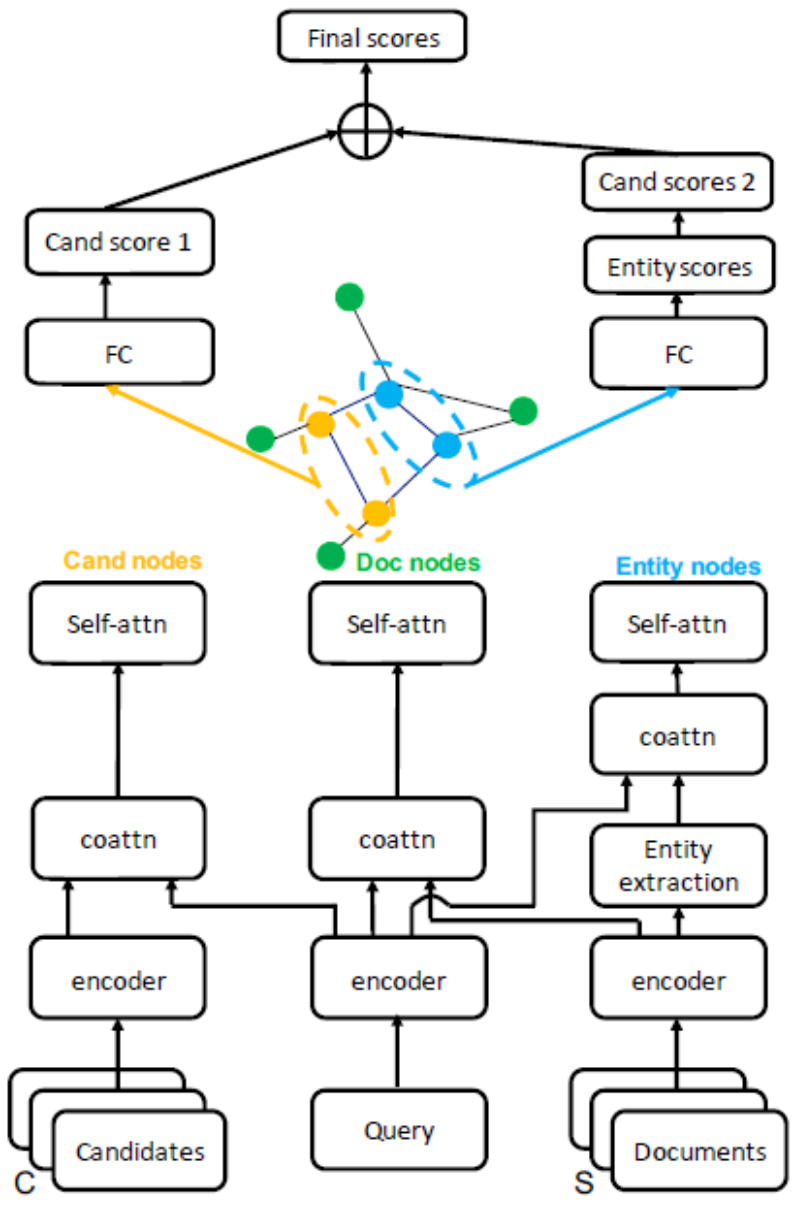

Figure 31: Architecture of HDE [46]

**SAE**: Tu et al. [47] proposed an interpretable system named Select, Answer, and Explain (SAE) with multi-hop reasoning graphs based on GNN with contextual sentence embeddings as nodes. This kind of sentence graph makes lead to an interpretable model because it can directly output supporting sentences with the answer prediction. The edges capture the global information presented within each document and also the cross-document reasoning path. Also, the contextual sentence embedding used in GNN is summarized over token representations based on a novel mixed attentive pooling mechanism. The attention weight is calculated from both answer span logits and self-attention output on token representations. This attention-based interaction enables the exploitation of complementary information between "answer" and "explain" tasks. The SAE system first filters unrelated documents, and selects gold documents using a document classifier trained with a novel pairwise learning-to-rank loss function. The gold documents are then fed into a model to predict the answer and supporting sentences. The model is a multi-task learning process which means while the answer prediction is accomplished at the token level, the support sentence is predicted as a node classification task at the sentence level. As it is shown in Figure 32, the selection module consists of:

- The *Multi-Head Self-Attention (MHSA)* layer to capture interaction among documents
- The *Pairwise Bi-Linear* layer to find gold documents accurately.

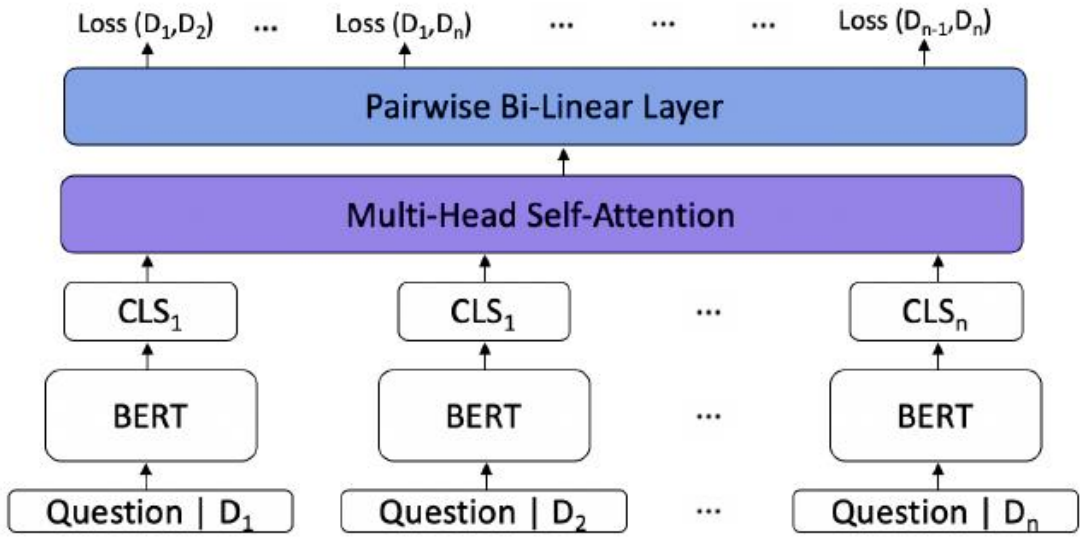

Figure 32: Architecture of the Selection module [47]

**HGN**: Fang et al. [48] presented a method named Hierarchical Graph Network (HGN) which uses different levels of granularity to construct the graph. In this graph, there are four types of nodes: questions, paragraphs, sentences, and entities to cover the different structures of the problem input. Besides, there are seven types of bidirectional edges including edges between the question node and paragraph nodes, edges between the question node and its entity nodes, edges between a paragraph node and its sentence nodes, edges between sentence nodes and their linked paragraph nodes, edges between a sentence node and its entities nodes, edges between paragraph nodes, and edges between sentence nodes from the same paragraph. As you can see in Figure 33 after constructing the hierarchical Graph, a graph-attention-based message passing algorithm is used for reasoning over the graph and finally, the multi-task prediction module is used for paragraph selection, supporting facts prediction, entity prediction, and answer span extraction. However, it uses link information from Wikipedia pages, which makes the graph inflexible for general use [14].

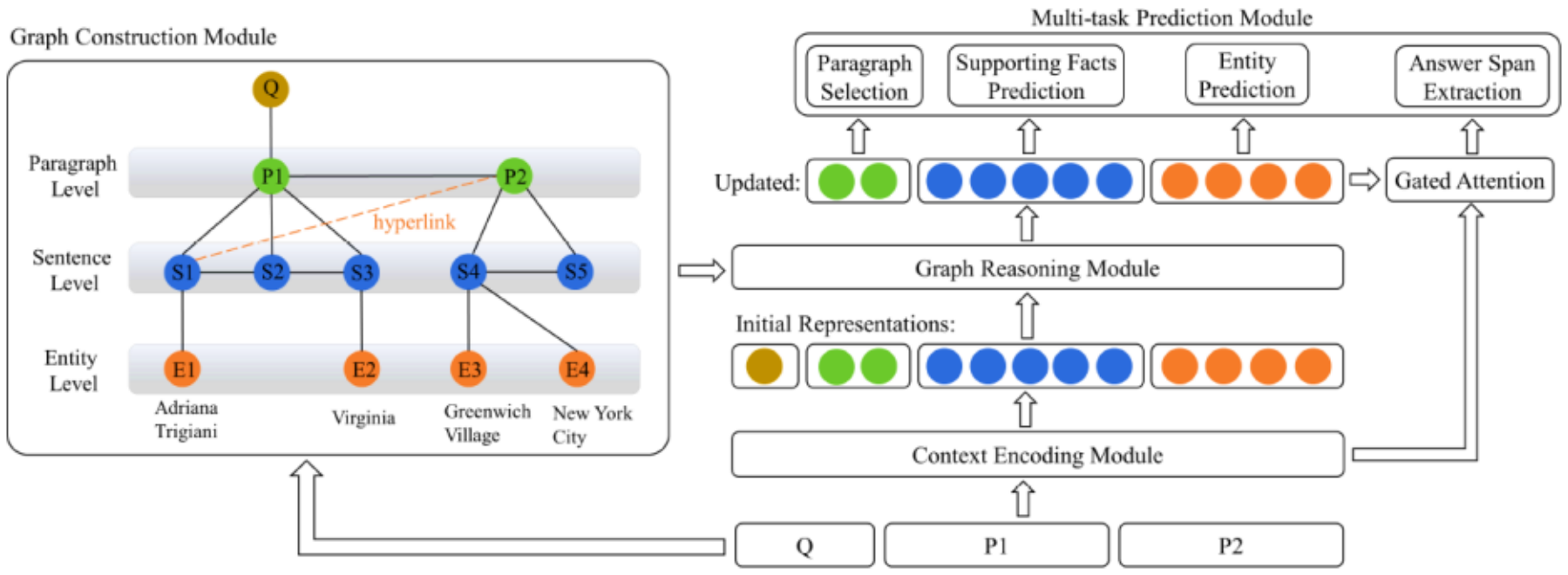

Figure 33: Architecture of HGN [48]

**HGNN**: Wang et al. [49] proposed a hierarchical graph neural network with a focus on compositional QA. In this study, the answers are derived from multiple but discontinuous segments in the documents. The nodes can be normal tokens, question tokens, sentence tokens, and special html image tokens. As you can see in figure (left), the input of the BERT sequence encoder is a question $q$, two sentences $(s_1 + s_2)$ and a special image node $h_1$ (some special tokens is used to indicate the question <SEP>, sentence <EOS> and special html image element <html>. The hierarchical graph neural networks blocks are shown in Figure 34 (middle). The attention-based Hierarchical Graph Neural Network (HGNN) uses three types of connections: 1) intra-sentence connection which means the connection of words within a sentence, 2) inter-sentence connection which means the connection of common tokens (e.g., sentence tokens or question tokens), and 3) global connection which means the connection between the question tokens and all the words in the document. The final prediction is made based on the sentence nodes and special html nodes. Finally, the connection mask matrix is used to connect the different tokens in the graph and predict the final answer.

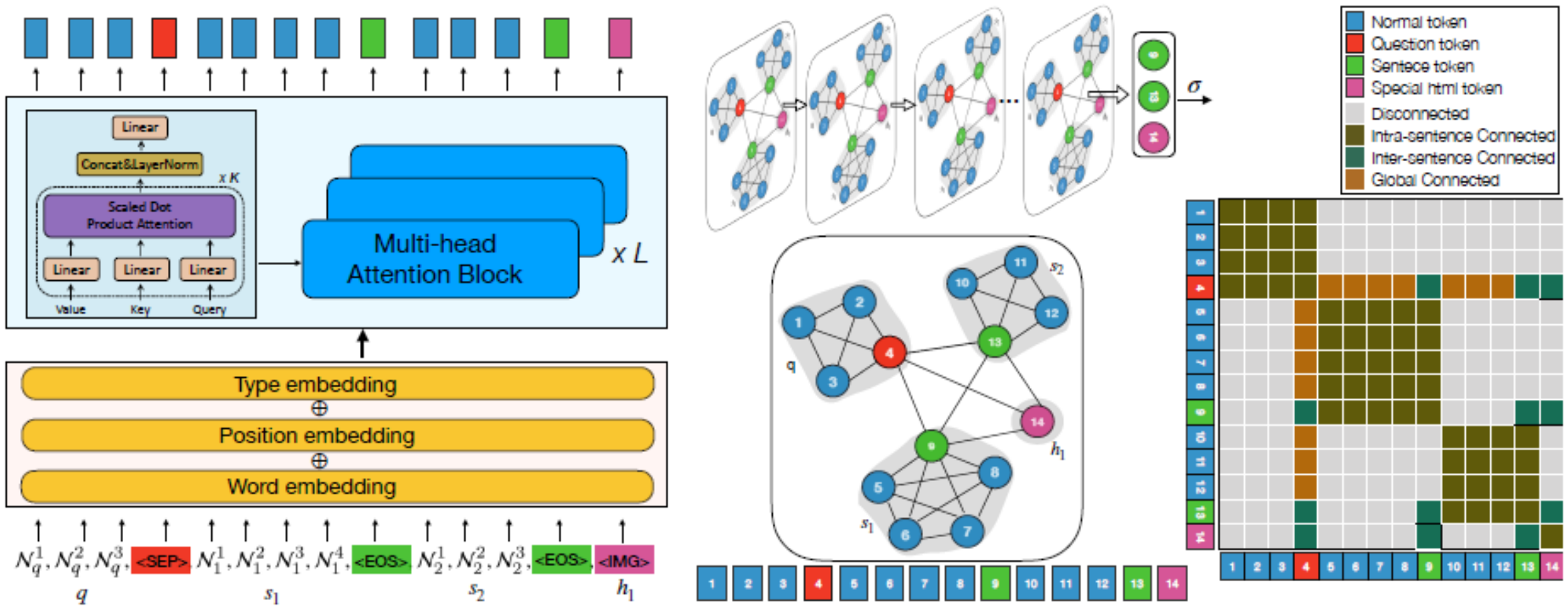

Figure 34: Architecture of HGNN [49]

**AMGN**: Li et al. [39] proposed a model named Asynchronous Multi-grained Graph Network (AMGN) for multi-hop MRC has been proposed. Previous studies perform message passing synchronously at each step of the graph update and ignore that different-level relationships have different priorities and the reasoning has to be done in an ordered logic. First, a multi-grained graph is constructed using the entity and sentence to reflect the relation level of the information. Second, an algorithm for asynchronous message propagation according to the relationship levels (e.g., entity-entity! entity-sentence! sentence-sentence) to update the graph to mimic human multi-hop reading logic is proposed. Besides, a question reformulation mechanism (RNN-based) is proposed to iteratively update the latent question representation with sentence nodes. These sentence nodes are directly used for supporting fact prediction. As it has been shown in Figure 35, the whole model consists of four main components: *Paragraph-selector* for reducing search space, *Encoder* to encode the context and question, *Construction & Reasoning* for multi-grained graph construction and multi-step asynchronous node update, and *Multi-task Prediction* to predict the final answer.

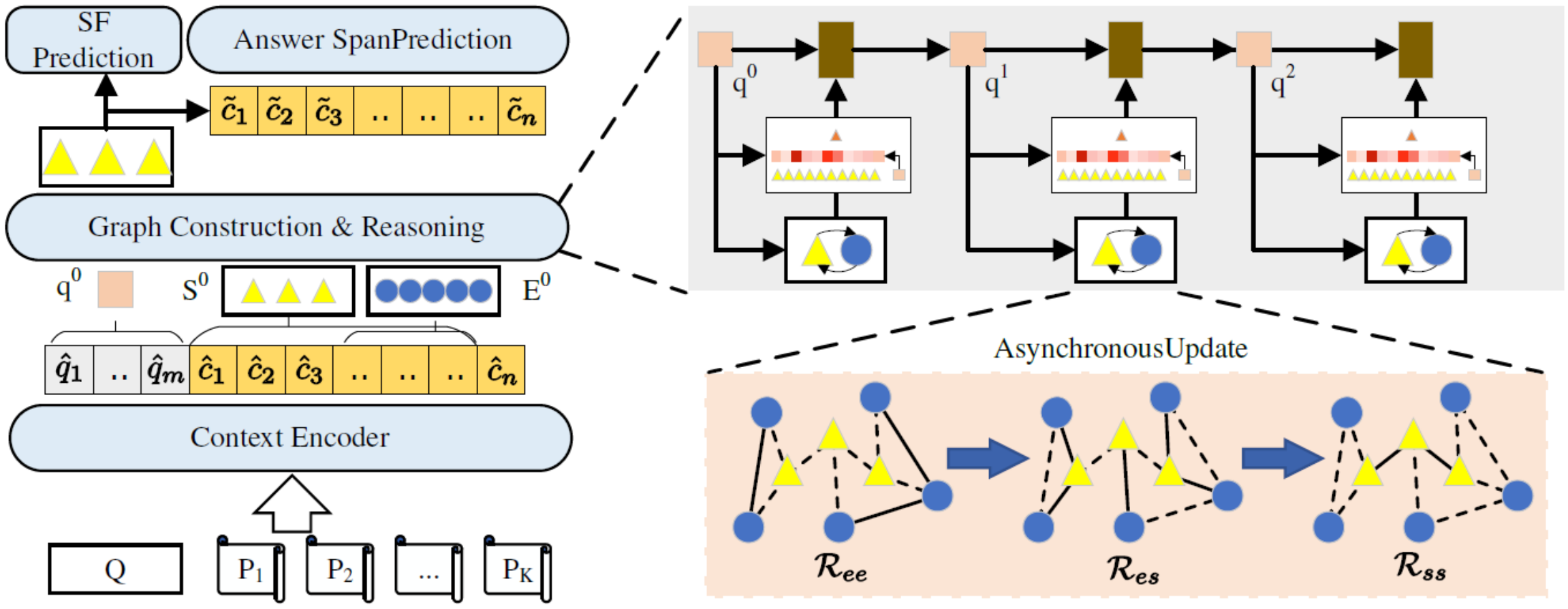

Figure 35: Architecture of AMGN [39]

**ClueReader**: Feng et al. [13] Proposes a model with a heterogenous graph and also based on the grandmother cells concept [9] to imitate the process of human brain. This concept explains that to answer a question, we generally recall a mountain of related evidence whatever the form it is (such as a paragraph, a short sentence, or a phrase), and coordinate theirs inter relationships before we carry out the final results. However, most of the studies on multi-hop MRC cannot gather the semantic features in multi-angle representations, which causes incomplete conclusions. Inspired by the concept of the Grandmother Cells in cognitive neuroscience, a spatial graph attention framework named *ClueReader* has been proposed. This model is designed to assemble the semantic features in multi-angle representations and automatically concentrate or alleviate the information for reasoning. The name "*ClueReader*" is a metaphor for the pattern of the model: assume the subjects of queries are the start points of clues, the reasoning entities are bridge points, and the latent candidate entities are the grandmother cells. the model has to achieve candidate entities from the clues. As you can see in Figure 36 after constructing a graph from different kinds of nodes, a GAT layer performs the message passing to find the final answer.

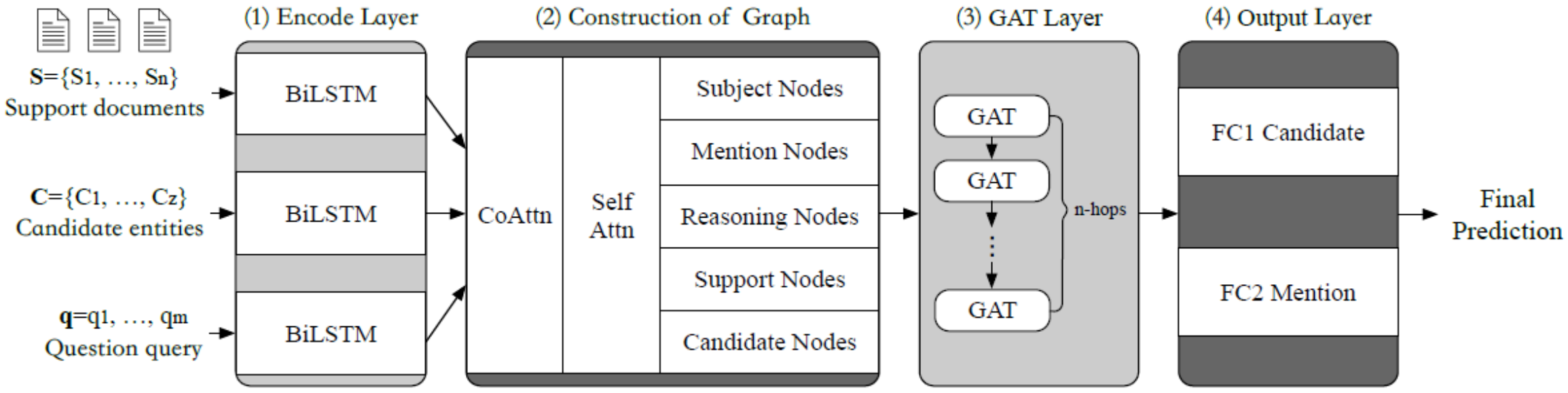

Figure 36: Architecture of ClueReader [13]

**IP-LQR**: Tang et al. [50] proposed a model named the latent query reformulation method (IP-LQR) to consider the phrase as nodes to construct the graph. As you can see in the example in Figure 37 when "Soviet statesman" is used as the reasoning starting point, there are three viable subsequent entities to follow, namely, "*Mikhail Gorbachev*", "*Nikolai Viktorovich Podgorny*"' and "*Andrei Pavlovich Kirilenko*". It can be non-trivial to choose from the three candidates to ensure that the reasoning process will eventually lead to the right answer. In contrast, if "former Soviet statesman" has been considered as the starting point, it will be much easier to locate the right subsequent entity "*Mikhail Gorbachev*" to update the query.

Then they proposed the latent query reformulation method (IP-LQR), which incorporates phrases in the latent query reformulation to improve the cognitive ability of the system. They also design a semantic-augmented fusion method based on the phrase graph, which is then used to propagate the information (Figure 38). After encoding the context and question into the high vector space and acquiring the representations of phrases, sentences, and paragraphs via the mean-pooling layer. Then, a similarity evaluation strategy is designed to calculate the weights of edges in the graph, also the fusion layer is used as an information aggregation to latently update the original question's representation. Finally, a re-attention mechanism is used to help locate the gold answer based on the new representation.

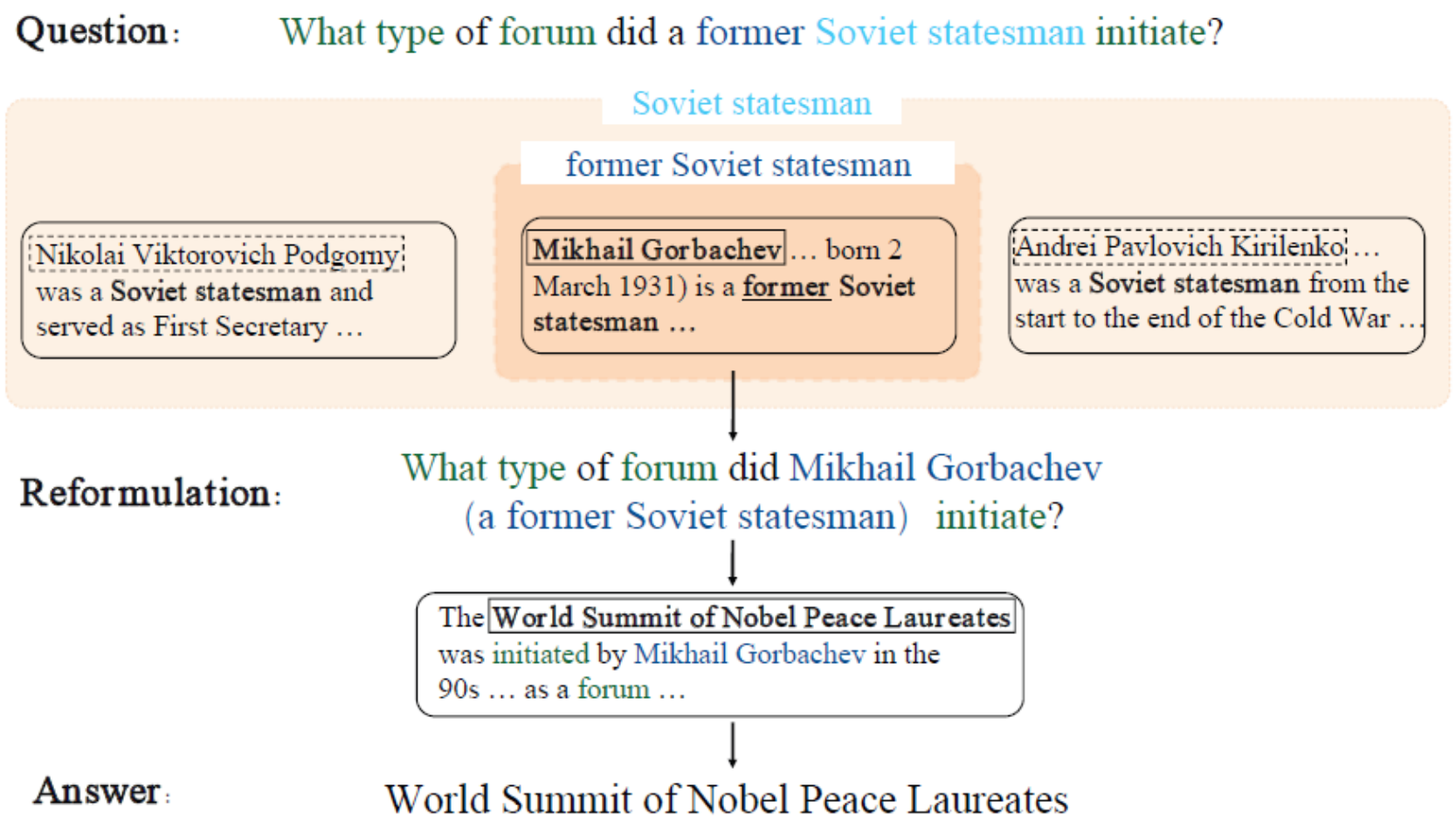

Figure 37: Example for the addressed challenges in IP-LQR [50]

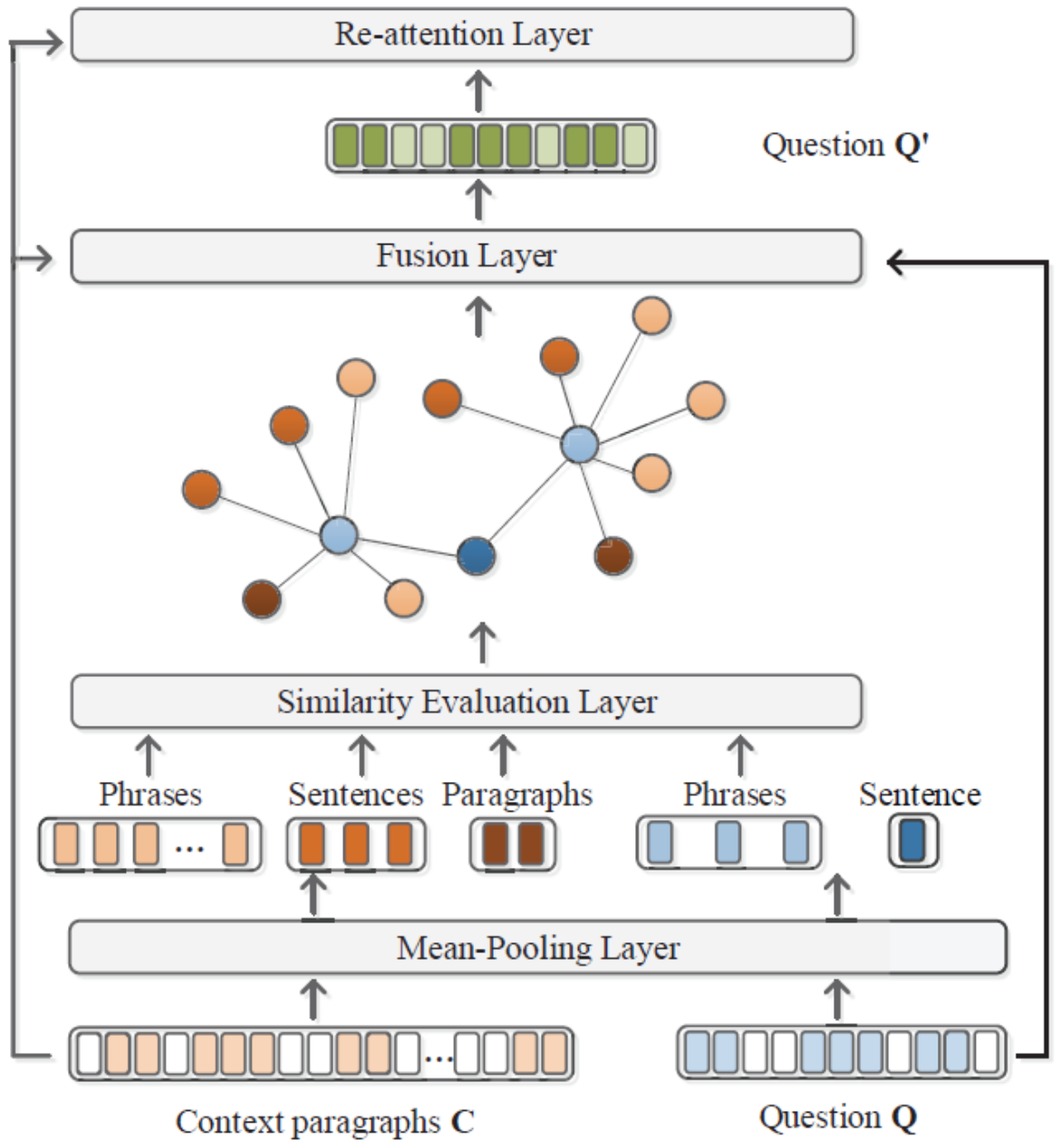

Figure 38: Architecture of IP-LQR [50]

**MKGN**: Ying et al. [51] proposed a model named Multi-Dimensional Knowledge Enhanced Graph Network for Multi-Hop Question and Answering (MKGN) utilizes dependency relations and commonsense knowledge for the reasoning process and proposed Multidimensional Knowledge enhanced Graph Network (MKGN). It uses specific knowledge to deal with information gaps and enhance representations of both questions and contexts in the reasoning process by using different dimensional knowledge, i.e., named entities, dependency relations and commonsense. There are two main components (Figure 39): 1*) Knowledge Extractor* extracts various knowledge from the contexts and the question and formulates them and 2) *Knowledge Enhancer* enhances representations of questions and contexts with each kind of knowledge generated by the knowledge extractor. Besides, they use the sequential and parallel manner kinds of fusion architectures. To stimulate a sequential reasoning process they fuse entity information, dependency relations, and commonsense one by one, also according to the fact that humans exploit multiple knowledge at the same time when making inferences they consider a parallel architecture for multi-dimensional knowledge utilization as well.

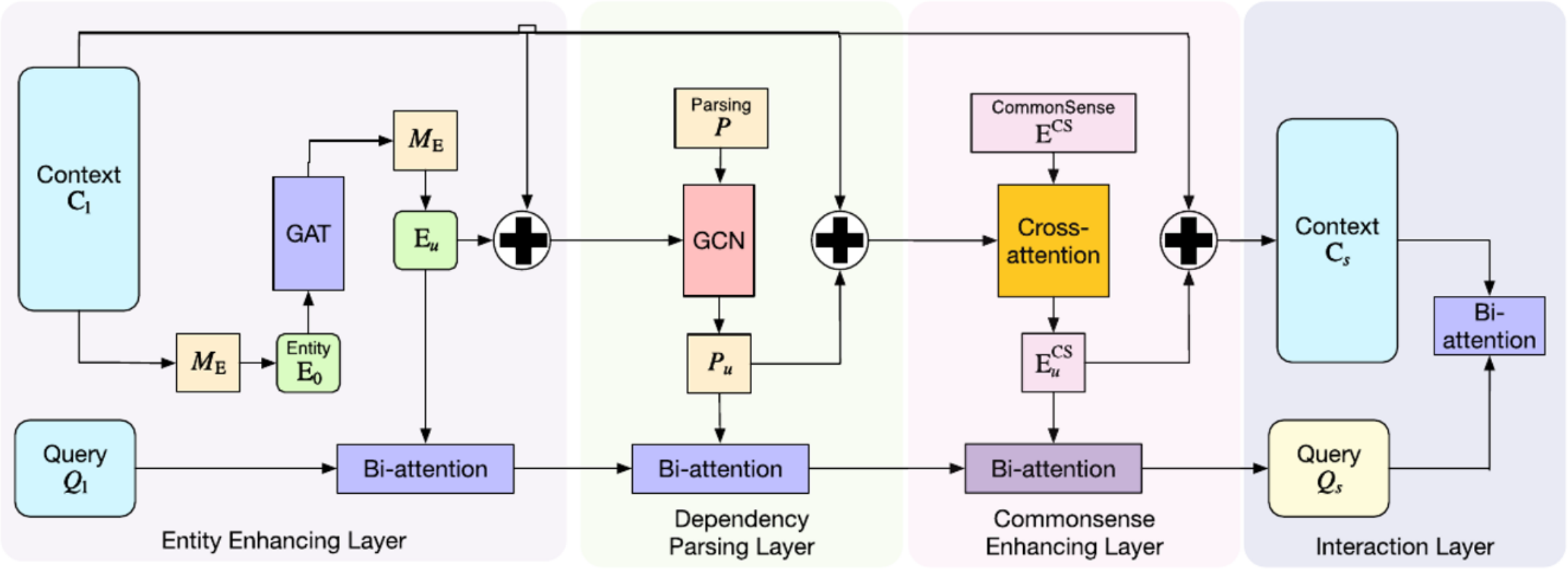

Figure 39: Architecture of MKGN [51]

Recently, graph-based models have become very popular and recognized as the main solution for multi-hop MRC because of the nature of modeling such a process into graph structure and the good results. But there is some drawback to this technique, the first one is the expensive computational process of the graph-based methods [52], and the second problem is that the graph-based models often can't cover all the inherent structure of documents and loss valuable structural information by modeling documents into graphs [12].

### 3.4 Graph-free technique

While graph-based methods were being dominated in multi-hop MRC, a question arose that whether the use of a graph is necessary despite the mentioned problem? Several studies addressed this important question, which we named the Graph-free technique because the main aim of them is to prove that graph is not necessary for multi-hop MRC. They may use other techniques to prove this fact (decomposition, recurrent reasoning or any other new techniques). Because of the importance of this question, we will review these studies in this section with a focus on the main idea of them.

**GF**: Shao et al. [53] attempted to answer this question: How much does graph structure contribute to answer a multi-hop question. They reimplement a graph-based model- Dynamically Fused Graph Network [54]- and claimed that the graph structure can play an important role only if the pre-trained models are used in a feature-based manner, while if the pre-trained models are used in the fine-tuning approach, the graph structure may not be helpful. Then the adjacency matrix based on manually defined rules and the graph structure can be regarded as prior knowledge, which could be learned by self-attention or Transformers. They proved that when texts have been modeled as an entity graph, both graph-attention and self-attention can achieve comparable results, but when texts have been modeled as a sequence structure, only a 2-layer Transformer could achieve similar results as DFGN. As you can see in Figure 40 when the entity graph is fully connected, a graph-attention layer will degenerate into a self-attention layer. However, this study as the first attempt of a graph-free model suffers from huge performance gap compared to the state-art-of the graph based-models [14].

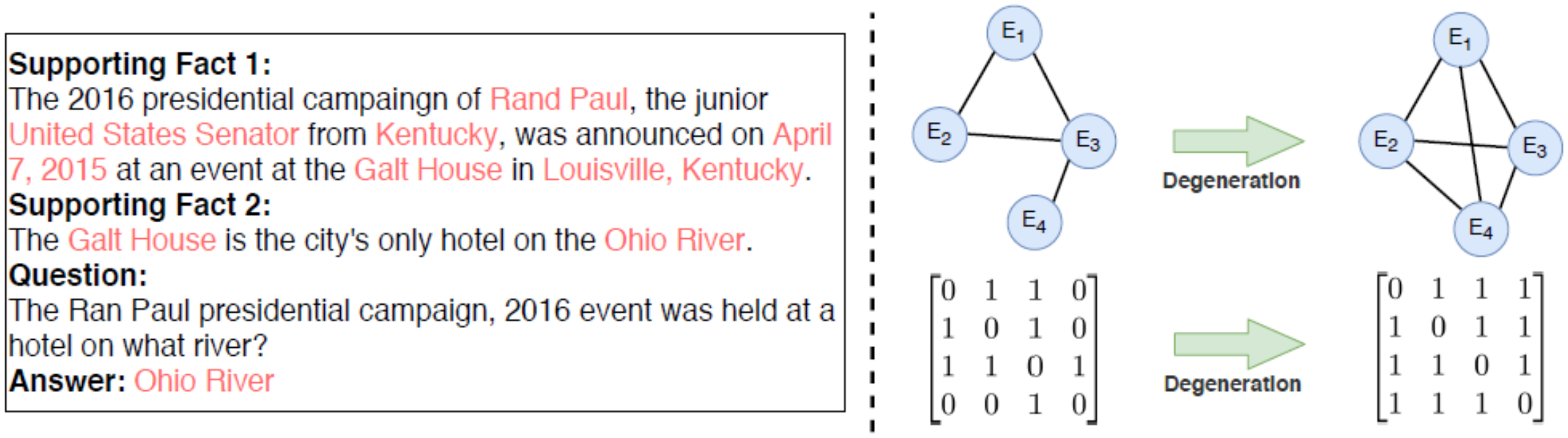

Figure 40: An example of the relation between the graph-attention network and the self-attention network [53]

**AMS**: Yuntao et al. [52] proposed another non-graph model with a focus on the document filters step to denoise irrelevant documents, and proves that if this step has been done properly, even a single-hop model can be used for multi-hop. They investigate that promising performance of the filter from Hierarchical Graph Network (HGN) (Fang et al., 2020) and showed that for 2-paragraph selection, both precision and recall can achieve around 95%, and for 4-paragraph selection, recall will be nearly 99%. Then they proposed Answer Multi-hop questions by Single-hop QA (AMS) models and used a single-hop QA models based on the attention mechanism with the HGN's document filter. As you can see in Figure 41 after the *Document denoise* layer, they used an *Attention-based single-hop* layer. AMS could achieve a comparable result with stat-of-the-art graph-bade models but still couldn't improve them.

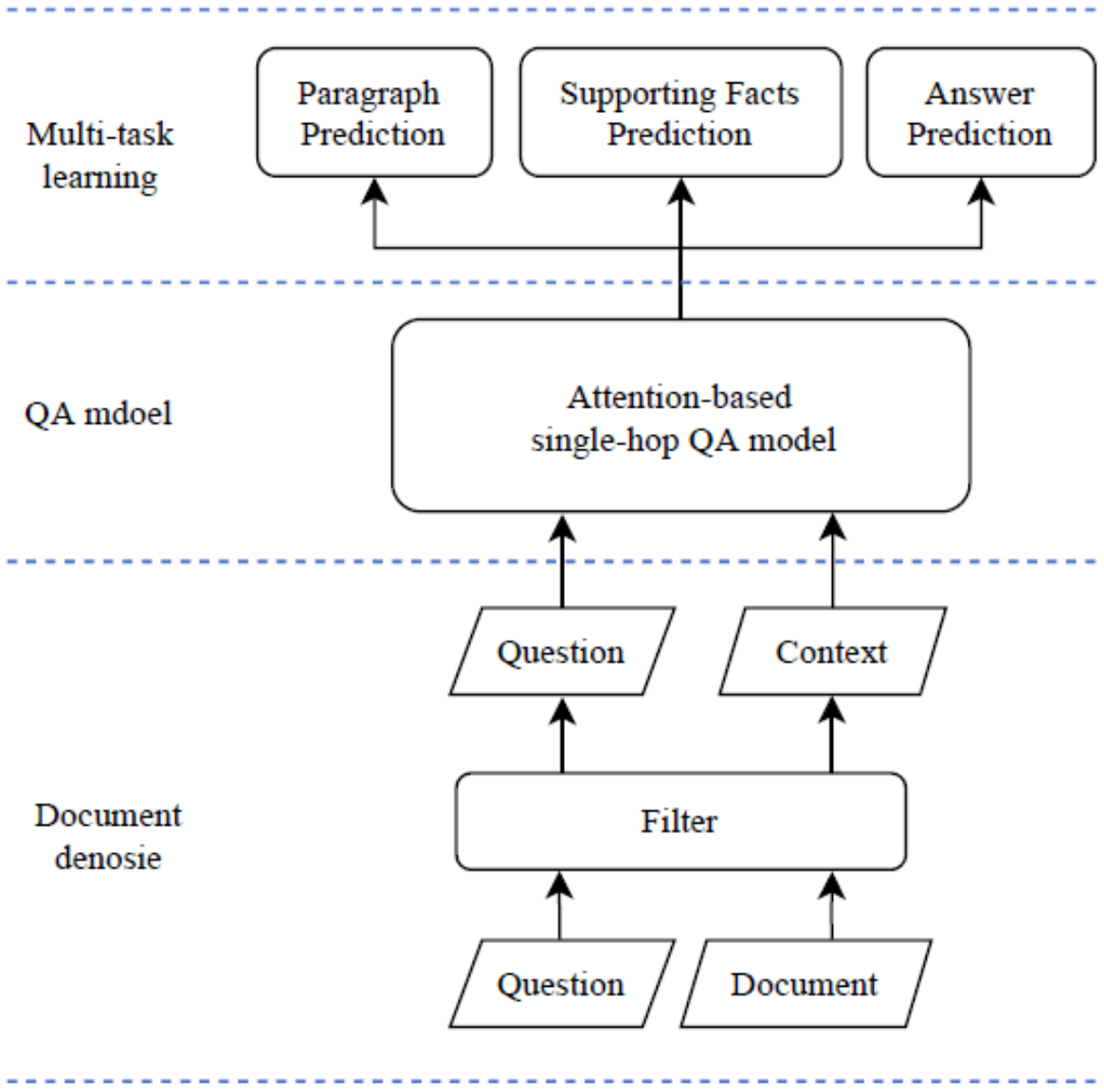

Figure 41: Overall architecture of AMS [52]

**S2G**: Wu, Zhang, and Zhao [14] investigated whether graph modeling is necessary for multi-hop. In this regard, they first proved that the retrieval stage is the most important module, while the existing studies focus on the reader module by graph modeling. This study presents a graph-free alternative named select-to-guide (S2G) to retrieve evidence paragraphs in a coarse-to-fine manner, incorporated with two attention mechanisms, which shows conforming to the nature of multi-hop reasoning. For the paragraph retrieval module, this study introduced a cascaded paragraph retrieval module that retrieves the evidence paragraphs in an explicit coarse-to-fine manner, and in multi-task module there are one shared encoder module alongside with two inter-dependent modules with an attention mechanism (Figure 42). However Concrete error analysis on S2G shows that there is still room for improvement on the multi-hop retriever module design.

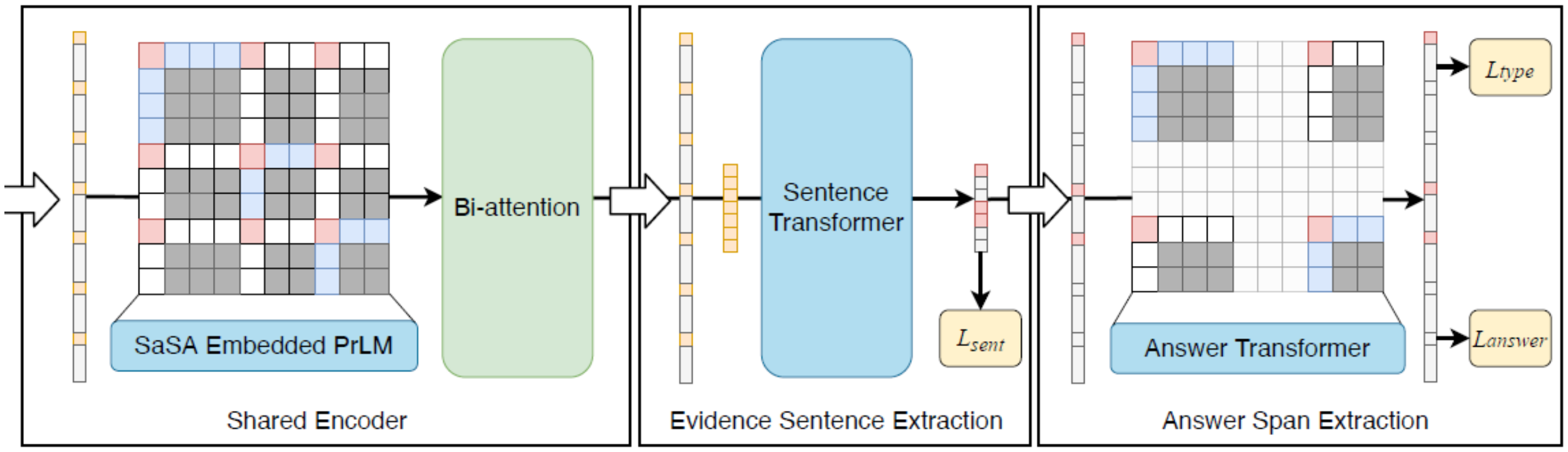

Figure 42: The multi-task module of S2G [14]

As the last part of this section, Figure 43 has been prepared to summarize the techniques and models.

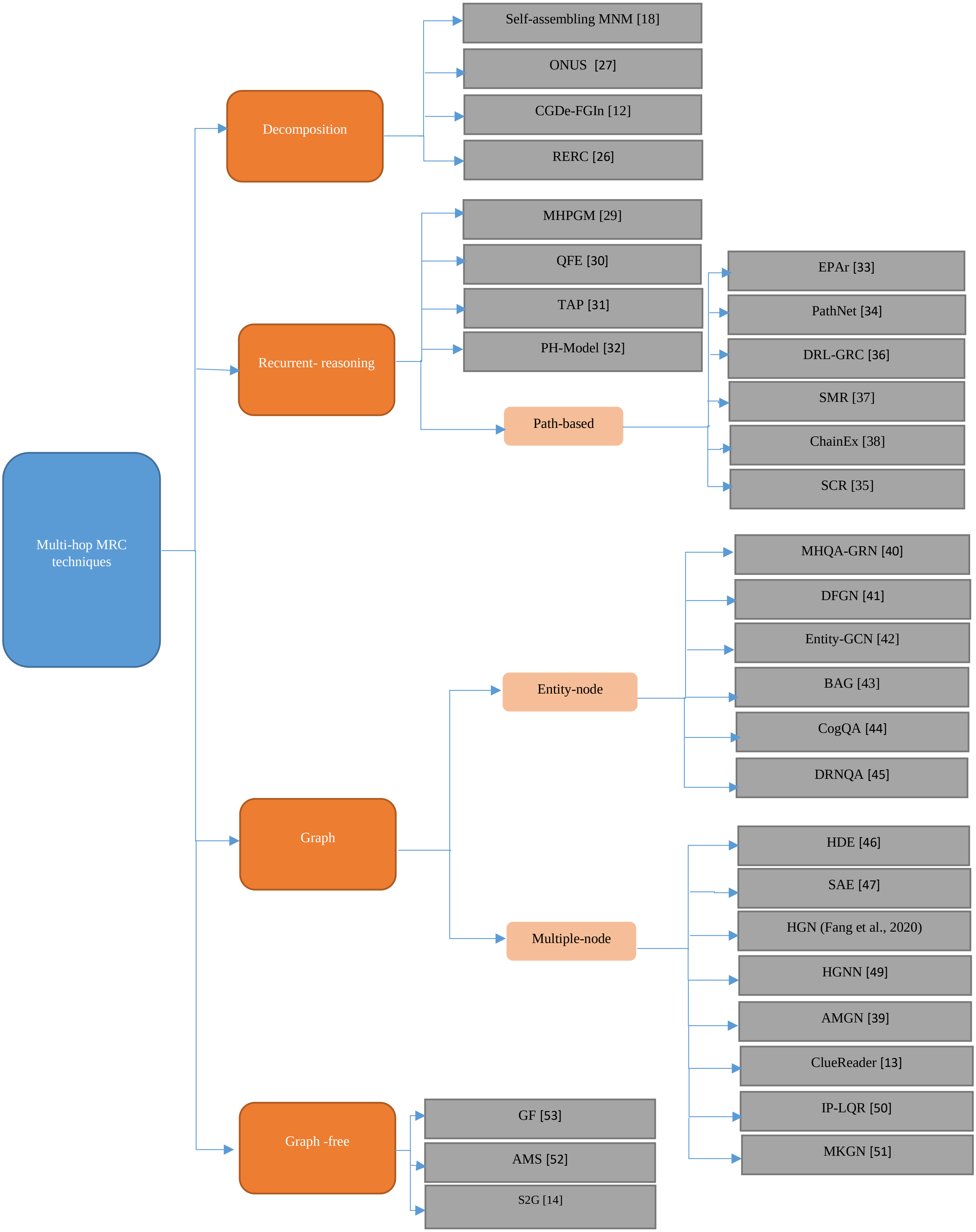

Figure 43: Multi-hop MRC techniques and models

## 4 COMPARISON

After reviewing the details of each model and categorizing them, in this section, a fine-grained comparison of models and techniques will be presented. In this regard, we first focus on the frequency of usage and popularity of each technique in recent years, then compare the performance of the models based on their final results.

### 4.1 Techniques frequency

The frequency of each technique among reviewed studies is shown in Figure 44. As you can see, the number of studies of the *Graph-based* techniques is the most, and after that the *Recurrent reasoning-based* technique has achieved good attention. But the number of studies cannot be enough to have a proper investigation, and it is needed to show the growth trend of each technique in different years. In this regard, Figure 45 shows the growth trend of each technique from 2018 to 2022. The graph-based technique has achieved the most attention in 2019, 2021, and 2022 that proves that popularity trend of this technique in different years as well. The first graph-free study has been proposed in 2020 and immediately this question was raised that whether the graph was really necessary due to the expensive computational? After that, some other studies followed this question and it can be said that can affect the popularity trend of the graph-based technique in future. However, graph-based technique still can be considered the most popular technique in multi-hop MRC.

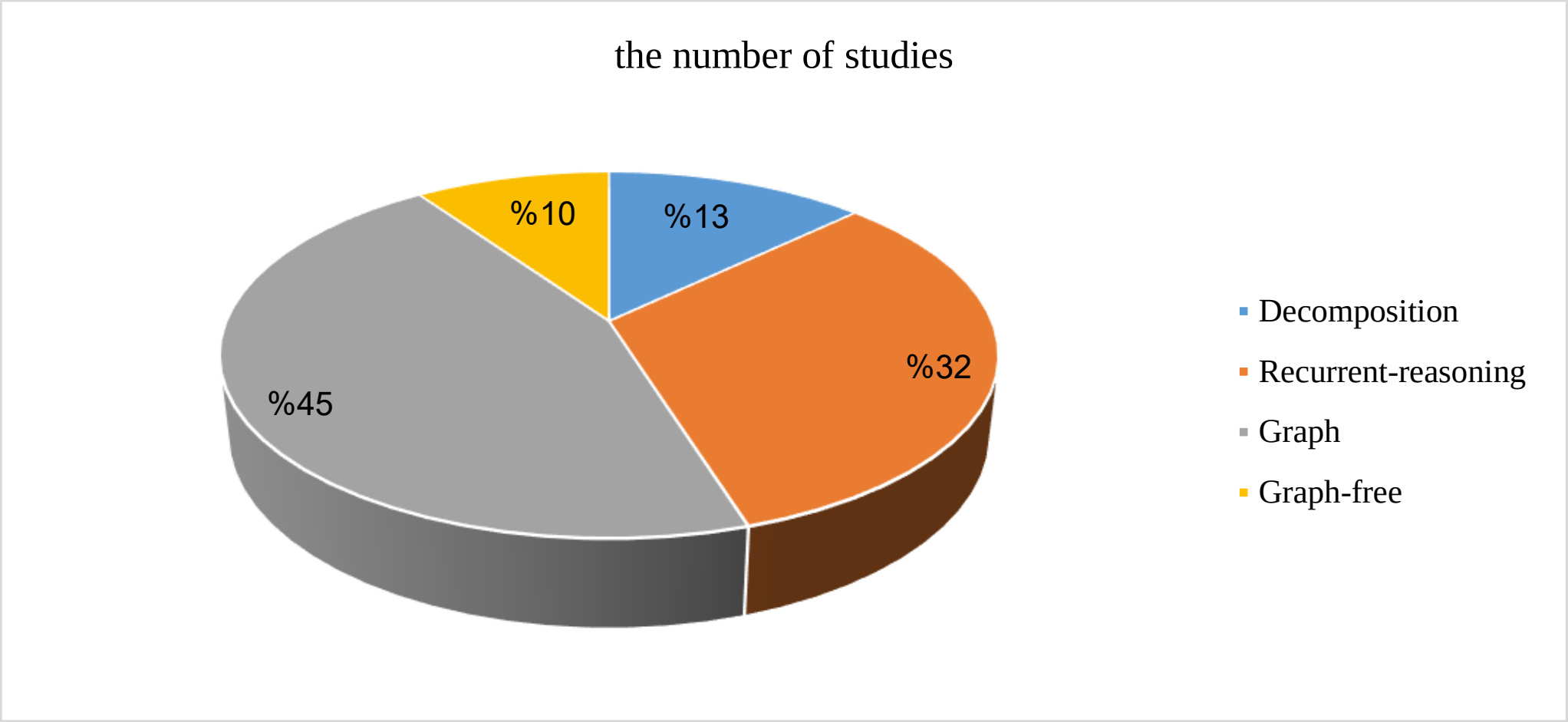

Figure 44: The number of studies in each technique

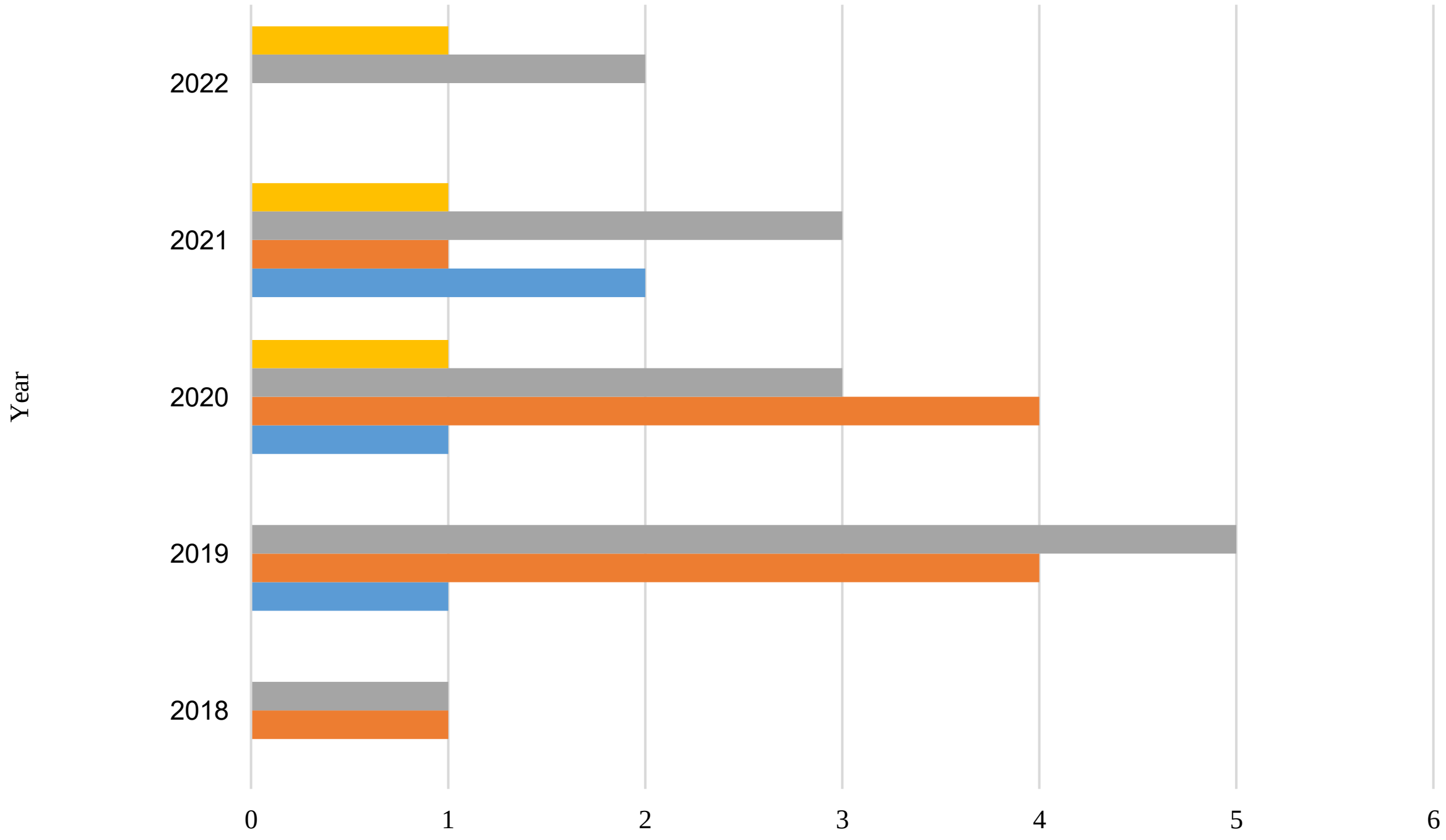

| | 2018 | 2019 | 2020 | 2021 | 2022 |
|---|---|---|---|---|---|
| Graph-free | 0 | 0 | 1 | 1 | 1 |
| Graph | 1 | 5 | 3 | 3 | 2 |
| Recurrent-reasoning | 1 | 4 | 4 | 1 | 0 |
| Decomposition | 0 | 1 | 1 | 2 | 0 |

Figure 45: The number of studies in each technique in recent years

### 4.2 Models Performance:

In this section we will show the performance of the models. This investigation is helpful in several ways; it will determine the stat-of-the-result, and also shows which models and techniques has achieved the best result. On the other hand, it can show the overall performance and effectiveness of each technique in multi-hop MRC

To evaluate the results of the models we need to use the evaluation metrics of the datasets. HotpotQA[11] and Wikihop[55], are two populare datasets among the reviewed studies as it has been clear in Figure 46 in which shows the percentage of use of two datasets among the reviewed models from 2018 to 2022. Then they provide a proper situation for evaluating the model's performance.

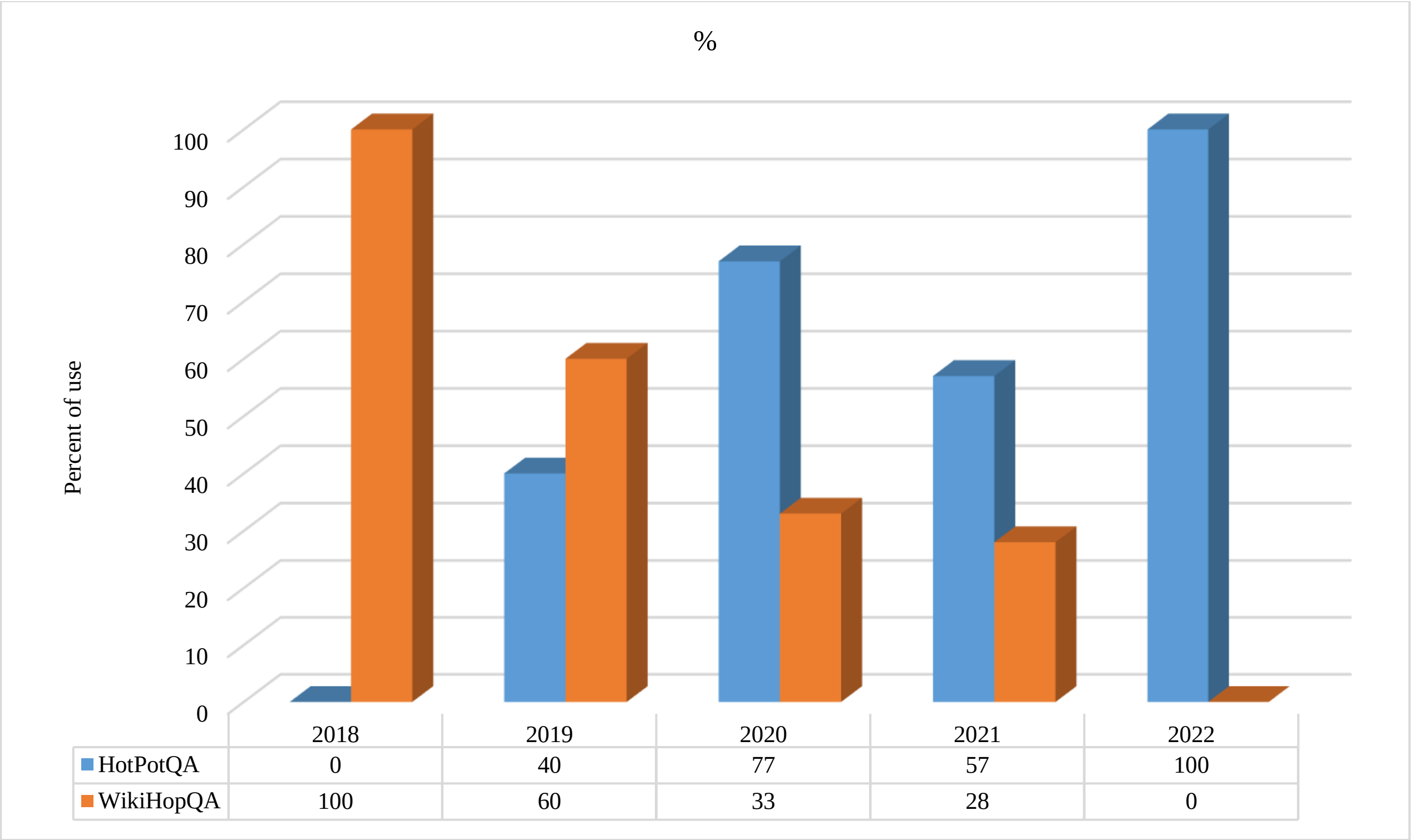

Figure 46: The percentage of use of HotpotQA and WikiHop among reviewed models

#### *4.2.1 HotpotQA*

In this section, the performance of models on HotpotQA [11] will be investigated. In this dataset, there are 90K training samples, 7.4K development samples, and 7.4K test samples. They have provided a question and 10 contexts in average, the corresponding answer, and supporting sentences to reach the answer. Thus, there are two sub-tasks in this dataset: Answer span prediction and Supporting facts prediction. Besides, there are two settings in HotpotQA: Distractor and Fullwiki. In the Distractor setting, for each question, two gold paragraphs with ground-truth answers and supporting facts have been provided. The Fullwiki setting is more challenging, which contains the same training samples as the Distractor setting, but it does not specify gold paragraphs for the test set to evaluate the model's ability in locating relevant facts.

Exact-match (EM) and F1 metrics have been used in this study to evaluate the model performance. EM is used to show whether the predicated answer and the ground-truth answer are exactly the same. F1 measures the average overlap between the predicted answer and the ground-truth answer. In addition, there are three sets of metrics: 1) Answer EM/F1 to evaluate the answer span prediction, 2) Support EM/F1 to evaluate the supporting facts prediction, and 3) Joint EM/F1 to combine the two previous ones.

To have an accurate comparison, the year of publications, the technique, along with the results, for Distracter setting and Fullwiki setting have been presented in this table 1 and 2. In the distractor setting, the best result is for **AMGN** [39] which is a graph-based model. Although the results are very diverse due to a large number of evaluation metrics, in general, graph-based methods have achieved relatively better results, and these good results were a motivation for great attention to the graph-based technique. But **S2G** [14] which is a graph-free model in 2021 has achieved a comparable result to the best graph-based model (Table 1). A few models have published the result for the Fullwiki setting due to the difficulty of this setting, and among them, **HGN** [48], which is a graph-based model, has achieved the best result (Table 2).

Table 1: Results of studies for the distractor setting of HotpotQA

| | | | Answer | | Support | | Joint | |
|---|---|---|---|---|---|---|---|---|
| Model | Year | Technique | EM | F1 | EM | F1 | EM | F1 |
| DFGN [41] | 2019 | Graph | 56.31 | 69.69 | 51.50 | 81.62 | 33.62 | 59.82 |
| Self-assembling MNM [18] | 2019 | Recurrent reasoning | 49.58 | 62.71 | | | | |
| QFE [30] | 2019 | Recurrent reasoning | 53.9 | 68.1 | 57.8 | 84.5 | 34.6 | 59.6 |
| HGN [48] | 2020 | Graph | 69.22 | 82.19 | 62.76 | 88.47 | 47.11 | 74.21 |
| ChainEX [38] | 2020 | Recurrent reasoning | 61.20 | 74.11 | | | | |
| SAE [47] | 2020 | Graph | 66.92 | 79.62 | 63.30 | 87.38 | 46.8 | 72.75 |
| DRNQA [45] | 2020 | Graph | 58.49 | 72.42 | 55.89 | 83.56 | 36.04 | 63.13 |
| TAP [31] | 2020 | Recurrent reasoning | 66.64 | 79.82 | 57.21 | 86.69 | 41.21 | 70.65 |
| CGDe-FGIn [12] | 2021 | Recurrent reasoning | 50.89 | 65.41 | 39.47 | 79.83 | 23.08 | 54.51 |
| **AMGN**[39] | 2021 | Graph | **83.37** | 83.46 | **88.83** | **89.13** | **75.24** | **75.48** |
| IP-LQR [50] | 2022 | Graph | 53.89 | 70.40 | 56.46 | 84.06 | 33.66 | 61.10 |
| GF [53] | 2020 | Graph-free | | | | | 44.67 | 72.73 |
| **S2G**[14] | 2021 | Graph-free | 70.72 | **83.53** | 64.30 | 88.72 | 48.60 | 75.45 |
| AMS[52] | 2022 | Graph-free | 68.87 | 82.14 | 63.20 | 88.45 | 46.67 | 74.21 |

Table 2: Results of papers for the FullWiki setting of HotpotQA

| | | | **Answer** | | **Support** | | **Joint** | |
|---|---|---|---|---|---|---|---|---|
| **Model** | **Year** | **Technique** | **EM** | **F1** | **EM** | **F1** | **EM** | **F1** |
| QFE [30] | 2019 | Recurrent reasoning | 28.7 | 38.1 | 14.2 | 44.4 | 8.69 | 23.1 |
| CogQA [44] | 2019 | Graph | 37.1 | 48.9 | 23.1 | 58.5 | 12.2 | 35.3 |
| **HGN** [48] | 2020 | Graph | **57.85** | **69.93** | **51.01** | **76.82** | **37.17** | **60.74** |

#### *4.2.2 WikiHop*

The papers that have used the WikiHop [55] dataset is investigated in this section. WikiHop consists of 51k questions, answers, and context where each context consists of several documents from Wikipedia .Each question in WikiHop is a tuple, which denotes two entities, and their relationship, then the answers in the WikiHop dataset are a single entity. Accuracy is a popular and fairly common metric to evaluate the performance of multiple-choice and Cloze-style MRC tasks. In the multiple-choice task, it is required to check whether the correct answer has been selected from the candidate answers. In contrast, in the Cloze-style task, it is required to check whether the correct words have been selected for the missing words

Since the answer type of this model is multiple-choice then accuracy is the evaluation metric on this dataset which is obtained for both the test and development set. For each paper, the year of publication, the technique along with the results are shown in Table 3. The best result is for **ChainEX** [38] which has used the recurrent reasoning technique. Besides that, the graph-based models could achieve the good result in this dataset too.

Table 3: Results of papers for WikiHop

| Model | Year | Technique | Accuracy (%) | |
|---|---|---|---|---|
| | | | Test set | Dev set |
| MHQA-GRN [40] | 2018 | Graph-based | 65.4 | 62.8 |
| MHPGM [29] | 2018 | Reasoning-based | 57.9 | 58.5 |
| DRL-GRC [36] | 2019 | Neural | --- | 65.12 |
| BAG [43] | 2019 | Graph-based | 69 | 66.5 |
| Entity-GCN [42] | 2019 | Graph-based | 71 | |
| **HDE** [46] | 2019 | Graph-based | **74.3** | **70.9** |
| PathNet [34] | 2019 | Reasoning-based | 69.6 | 67.4 |
| EPAr [33] | 2019 | Reasoning-based | 69.1 | 67.2 |
| **ChainEX** [38] | 2020 | Graph-based | **76.5** | **72.2** |
| SMR [37] | 2020 | Reasoning-based | 68.3 | --- |
| SCR[35] | 2020 | | | 71.6 |
| Heterogenous[13] | 2021 | Graph | 72.0 | 66.5 |

# 5 OPEN ISSUES

In this section, some open issues in multi-hop MRC will be discussed.

The lack of free-form and close-style datasets has made it impossible to propose models for these two tasks, while both tasks have a good potential to use in form of multi-hop MRC. Also, among all the types of MRC tasks described in Section 2, the free-form task is the most complex type of MRC, and due to its complexity, only a few studies have been focused on it. Since this type of answer is more similar to real-world scenarios, focusing on presenting more Free-form datasets and models can improve the application of MRC systems.

# 6 CONCLUSION

In this study, we focused on the multi-hop MRC approaches. In this regard, after presenting the multi-hop MRC problem definition, the multi-hop MRC techniques had been explained based on 31 studies from 2018 to 2022. In addition to categorize the approaches based on the main technique, they also were reviewed in detail including the architecture, superiority, and motivations. In the following, a fine-grain comprehension of the approaches and techniques was prepared, and finally, some open issues in this field were discussed.